\documentclass{article}

\usepackage[final]{nejlt}

\usepackage{lipsum}
\usepackage{graphicx}
\usepackage{newfloat}
\usepackage{listings}
\usepackage{amsmath}
\usepackage{multirow}
\usepackage{subcaption}
\usepackage{algorithm}
\usepackage{algorithmic}

\title{Unsupervised Text Embedding Space Generation Using Generative Adversarial Networks for Text Synthesis}

\author{
  Jun-Min Lee, Korea Advanced Institute of Science and Technology, I-BRICKS, {\tt \small ljm56897@gmail.com} \and
  Tae-Bin Ha, I-BRICKS, {\tt \small taebinalive@gmail.com}
}

\begin{document}
\thispagestyle{title}

\abstract{
Generative Adversarial Network (GAN) is a data synthesis model that creates plausible data through the competition between a generator and a discriminator.  
Although GAN has been extensively studied for image synthesis, it has inherent limitations when applied to natural language generation.
This is because natural language is composed of discrete tokens, and the generator faces challenges in updating its gradient through backpropagation.
Therefore, most text-GAN studies generate sentences starting with a random token (or prompt) based on a reward system.
Thus, the generators of previous studies are pre-trained in an autoregressive manner before adversarial training, resulting in data memorization where synthesized sentences reproduce the training data.
In this paper, we synthesize sentences using a framework similar to the original GAN.
More specifically, we propose Text Embedding Space Generative Adversarial Networks (TESGAN), which generate continuous text embedding spaces instead of discrete tokens to address the gradient backpropagation problem.
Furthermore, TESGAN conducts unsupervised learning that does not directly refer to the text of the training data to overcome the data memorization issue.
Also, TESGAN enables unconditional text synthesis during the inference phase by using random noise instead of tokens or prompts for text synthesis.
By adopting this novel method, TESGAN can synthesize new sentences, demonstrating the potential of unsupervised learning for text synthesis.
We look forward to extended research that combines large-scale language models with a new perspective on viewing text as continuous spaces.
}

\maketitle

\section{Introduction}
Generative Adversarial Network (GAN), as proposed by \citet{NIPS2014_5ca3e9b1}, is a popular model for data synthesis.
GAN is an unconditional data generation algorithm that aims to generate plausible data in an unsupervised manner by fostering competition between a generator and a discriminator to capture the real data distribution.
When GAN was initially introduced, it primarily focused on image synthesis, and extensive research was conducted to achieve high-quality synthetic data results~\cite{pmlr-v70-arjovsky17a,https://doi.org/10.48550/arxiv.1511.06434,https://doi.org/10.48550/arxiv.1812.04948,https://doi.org/10.48550/arxiv.2106.12423}.
Furthermore, GAN is commonly employed in the field of computer vision for data augmentation through image synthesis~\cite{article,bowles2018gan,antoniou2018data,10.1109/TIP.2021.3049346}.
The GAN generator learns implicit density based on the discriminator's loss without direct reference to the training data.
Consequently, GAN can prevent data memorization, where the model reproduces the training data.
Additionally, GAN can synthesize various data by using random noise instead of a specific starting point, such as a designated start token.

Similar to images, unconditional text generation can function as a data augmentation technique by generating new text that resembles a given dataset.
It also has practical applications, such as creating new documents by generating fictitious text information suitable for direct use. 
Consequently, several studies have attempted to apply GAN to natural language, but they have encountered limitations in natural language generation.
The challenge arises from the fact that natural language is composed of discrete tokens, making it challenging for the GAN generator to directly update gradients through backpropagation.
The gradient backpropagation issue in text-based GANs was first discussed by \citet{Yu_Zhang_Wang_Yu_2017}, and numerous subsequent text-GAN research efforts aimed to address this problem using gradient policy-based reinforcement learning with a reward system.
Furthermore, the previous text-GAN approaches necessitated pre-training the generator with supervised learning (autoregressive) before adversarial training due to convergence issue with the generator \cite{Yu_Zhang_Wang_Yu_2017}.
Accordingly, we discovered that the generators of previous text-GAN approaches reproduce the training data (leading to data memorization) during text synthesis due to the autoregressive-based pre-training process, which becomes a significant issue in generative models.

This paper introduces a novel framework known as Text Embedding Space Generative Adversarial Networks (TESGAN)\footnote{\url{https://github.com/ljm565/TESGAN}}, which enables backpropagation and prevents data memorization.
TESGAN does not rely on a supervised, pre-trained autoregressive-based generator that generates discrete tokens for text synthesis.
Our generator generates continuous text embedding spaces for text synthesis instead of discrete tokens, allowing training with gradient backpropagation.
Furthermore, the fact that TESGAN deals with continuous spaces makes it possible for TESGAN's generator to be trained within the original GAN framework to mimic the real text embedding space.
Moreover, TESGAN enables unconditional text generation, as it does not require the selection of a starting token (or prompt) for text synthesis.
Our seed interpretation model then synthesizes sentences by interpreting the imitated continuous text embedding space created by the generator.
During sentence synthesis, data memorization does not occur because TESGAN does not directly refer to the training text data but only learns from the continuous text embedding space.
We use two datasets to conduct performance evaluations and general applicability experiments based on synthetic text generated by TESGAN.
To assess the quality and diversity of synthesized text, we employ evaluation metrics such as Fréchet BERT Distance (FBD), Multi-sets-Jaccard (MSJ)~\cite{alihosseini-etal-2019-jointly}, Language Model score (LM)~\cite{dAutume2019TrainingLG,DBLP:conf/iclr/CacciaCFLPC20}, and Self-BLEU (SBL)~\cite{10.1145/3209978.3210080}.
In addition, we conducted human evaluations, and TESGAN achieved the highest average score.
Lastly, we calculate the data memorization ratio and present the synthesized sentences to assess the potential of unsupervised learning and continuous embedding spaces for text synthesis.

\section{Related Works}
The most common method of text generation is to use an autoregressive-based language model via teacher forcing~\cite{6795228}.
For example, extensive studies have been conducted on models using recurrent neural network (RNN) with Long Short-Term Memory (LSTM) cells~\cite{10.1162/neco.1997.9.8.1735}.
Using LSTM, \citet{DBLP:journals/corr/Graves13} successfully generated handwriting by predicting sequences, and \citet{wen-etal-2015-semantically} synthesized sentences under specific conditions.
\citet{bowman-etal-2016-generating} generated text after learning text embedding spaces with an autoregressive-based LSTM model and a variational autoencoder (VAE) architecture \cite{Kingma2014}.
Policy Gradient with BLEU (PG-BLEU) calculates the BLEU~\cite{papineni-etal-2002-bleu} score of synthesized sentences and takes them as a reward when updating the generator using policy gradient.

Numerous investigations have been conducted to utilize GANs for text synthesis.
Sequence GAN (SeqGAN)~\cite{Yu_Zhang_Wang_Yu_2017} attempted to address the backpropagation problem by employing gradient policy-based reinforcement learning with a reward system. 
However, SeqGAN faced a reward sparsity issue, leading \citet{10.5555/3294996.3295075} to introduce RankGAN, which replaced the previous regression-based discriminator with a novel ranker.
RankGAN trains the discriminator to assign higher scores to more realistic sentences.
MaskGAN \cite{DBLP:conf/iclr/FedusGD18} utilized an LSTM-based generator to fill in masked parts of sentences with tokens during training.
Since MaskGAN uses discrete tokens, gradient backpropagation is not possible for the generator.
To overcome this challenge, the authors employed the actor-critic method, using the probabilities of candidate tokens from the discriminator as rewards during training.
\citet{https://doi.org/10.48550/arxiv.1702.07983} proposed Maximum Likelihood Augmented Discrete GAN (MaliGAN), which synthesizes text by minimizing Kullback-Leibler divergence~\cite{10.1214/aoms/1177729694}.
LeakGAN~\cite{guo2018long} alleviated issues related to sparseness and the lack of intermediate information by providing leaked information from the discriminator.

Several studies have aimed to address the gradient backpropagation problem without relying on reward-based reinforcement learning.
TextGAN~\cite{zhang2017adversarial} introduced kernel-based moment-matching, which enforces empirical distributions of real and synthetic text by using LSTM and Convolutional Neural Networks (CNN) for the generator and the discriminator, respectively. 
Feature Mover GAN (FM-GAN)~\cite{10.5555/3327345.3327377} defined the feature-mover's distance (FMD) and learned it by minimizing the FMD between real and fake sentences.
Both TextGAN and FM-GAN utilized LSTM generators that generate discrete tokens using the \textit{soft-argmax} trick instead of relying on reinforcement learning.
Relational GAN (RelGAN)~\cite{DBLP:conf/iclr/NieNP19} applied relational recurrent neural networks~\cite{10.5555/3327757.3327832} and attempted to address the gradient backpropagation issue using Gumbel-softmax~\cite{DBLP:conf/iclr/JangGP17}.
However, since these approaches employed autoregressive (e.g., LSTM) generators, they explicitly referenced the training text data during model training.
Consequently, previous studies faced challenges in avoiding complete data memorization while synthesizing sentences due to an autoregressive generator.
Lastly, Transformer-based Implicit Latent GAN (TILGAN)~\cite{DSSSZ2021} adopted a similar approach to TESGAN for addressing the gradient backpropagation issue based on the embedding space. 
However, TILGAN differs from TESGAN in that it was trained on a latent space compressed by the encoder, configured as an autoencoder transformer, and did not utilize embeddings learned from real language models.

Most of the aforementioned text-GAN models require the first token or prompt to synthesize text due to their autoregressive generators.
TESGAN stands apart from these models as it generates text embedding space directly from random noise, eliminating the need for selecting tokens.
Our TESGAN is the first text-GAN model that learns the real text embedding space without relying on an autoregressive generator.

\section{Text Embedding Space GAN}
TESGAN aims to generate the seeds required for synthesizing plausible text.
These generated seeds (fake seeds) from the generator, along with the real seeds from the real text, are passed to the discriminators for training within the GAN framework.
Once the training of TESGAN is complete, the pre-trained seed interpretation model synthesizes text using the fake seed created by the generator.
\begin{figure*}[t]
    \centering
    \includegraphics[width=1\textwidth]{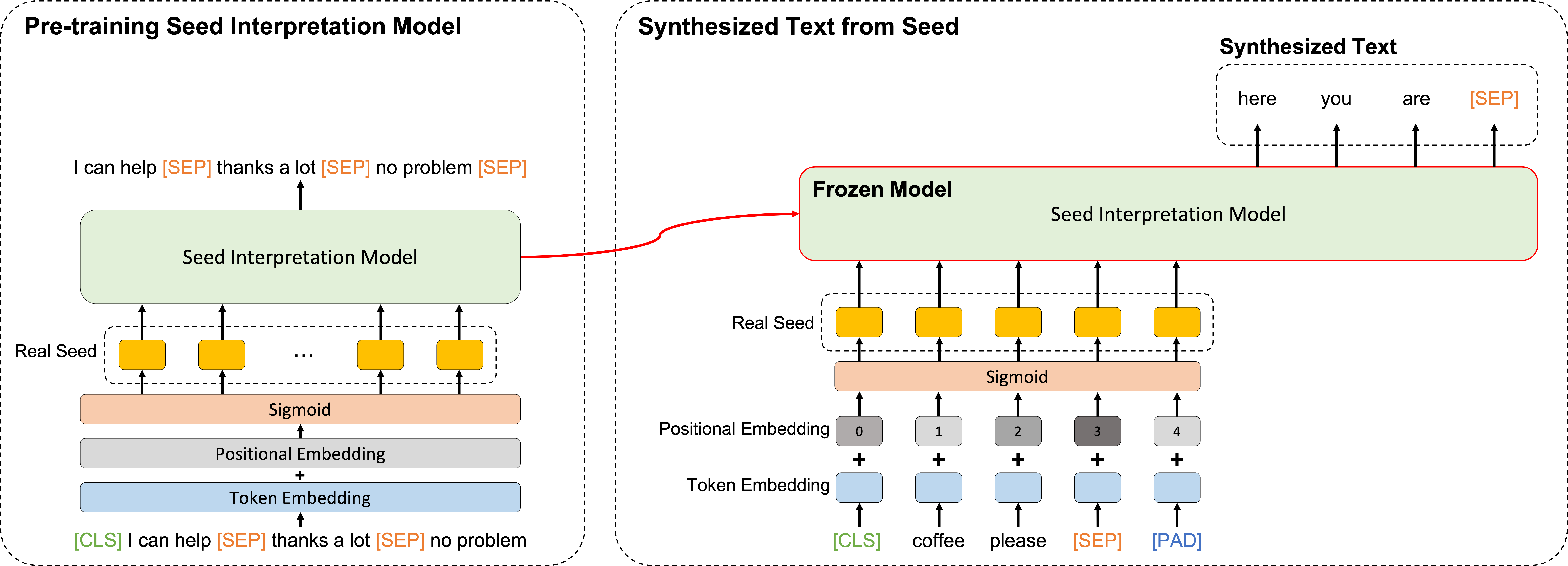}
    \caption{Illustration of the seed interpretation model.
            The seed interpretation model is pre-trained with multi-turn sentences before adversarial training (left).
            After pre-training, the model's parameters are frozen, allowing it to synthesize text from the seed. The right figure implies that text can be synthesized from the seed.
            The $[PAD]$ tokens following the $[SEP]$ tokens are omitted in the left part for clarity.}
    \label{seedInterpretationModel}
\end{figure*}

\subsection{Seed for Text Synthesis}
\label{seed}
We denote a text sequence as $S=w_{1}, \ldots, w_{T}$ ($T$ is the sequence length).
An autoregressive-based language model calculates the probability of the text sequence $S$ as a product of conditional probabilities.
If we assume that $S$ is a complete sentence, then the sequence $D=S_{1}, \ldots ,S_{N}$ ($N$ is the dialogue length) can be viewed as multi-turn sentences.
Let $S_{1}=w_{1}, \ldots ,w_{m}$ and $S_{2}, \ldots ,S_{N}=w_{m+1}, \ldots ,w_{M}$ denote the first sentence and the subsequent sentences, respectively ($M$ is the total length of the multi-turn sentences $D$). 
The subsequent sentences after the first sentence can be predicted from a product of conditional probabilities:
\begin{equation}
    \small
    p(S_{2}, \ldots, S_{N}|S_{1})=\prod_{i=m+1}^{M}p(w_{i}|w_{1}, \ldots ,w_{i-1})
    \label{equation1}
\end{equation}
In other words, the first sentence can generate subsequent text using an autoregressive-based language model trained with multi-turn sentences.
Therefore, the first sentence can serve as a seed.
Meanwhile, the generator of TESGAN generates the first sentence as a continuous embedding space instead of discrete tokens to enable gradient backpropagation.
Consequently, a continuous embedding space of the first sentence is defined as a seed.

\subsection{Seed Interpretation Model}
We define a seed in Section~\ref{seed}, and the seed interpretation model $f_{\theta}(\cdot)$ is used to synthesize text based on the seed.
To synthesize text, the seed interpretation model must first be trained with multi-turn sentences in an autoregressive manner, similar to general language modeling, before adversarial training, as shown in Figure~\ref{seedInterpretationModel} (left), with the following loss function:
\begin{equation}
  \mathcal{L}_{LM}=-\frac{1}{N}\sum^{N}_{n=1}{\log{\frac{exp(x_{n, y_n})}{\sum^{C}_{c=1}{exp(x_{n,c})}}}}
  \label{lmloss}
\end{equation}
This enables the generator to synthesize appropriate text by utilizing the fake embedding space it creates during the inference phase.
More detailed explanations will be provided in the following section.
As a result, the model has to be trained on data consisting of multi-turn sentences $D=S_{1}, \ldots ,S_{N}$, where each sentence has a maximum length of $L$, meaning the total number of tokens in $D$ is $N\times L$.
When constructing the multi-turn sentence data, the special token $[CLS]$ is inserted only at the beginning of the first sentence, and each sentence is distinguished by adding the special token $[SEP]$ at the end.
If the length of the tokenized sentence is less than $L$, the sentence is padded with the special token $[PAD]$:
\begin{equation}
    \small
    \begin{gathered}
        S_{1}=w_{1}^{1}, \ldots , w_{|S_{1}|}^{1}, \ldots , w_{L}^{1} \\
        (w_{1}^{1}=[CLS], w_{|S_{1}|}^{1}=[SEP], w_{|l>S_{1}|}^{1}=[PAD])
    \end{gathered}
    \label{equation2}
\end{equation}
\begin{equation}
    \small
    \begin{gathered}
        S_{i(i>1)}=w_{1}^{i}, \ldots , w_{|S_{i}|}^{i}, \ldots , w_{L}^{i} \\
        (w_{|S_{i}|}^{i}=[SEP], w_{|l>S_{i}|}^{i}=[PAD])
    \end{gathered}
    \label{equation3}
\end{equation}
where $S_1$ and $S_i$ represent a seed sentence and subsequent text.
Let $H_{real}$ denote the real seeds from TESGAN.
As shown in Figure~\ref{seedInterpretationModel}, the real seed is an embedding space of a sentence obtained by applying the sum of the token embedding and the positional embedding to the sigmoid function.
Since most sentences can exist before others as long as the seed interpretation model is trained with multi-turn sentences, a significant portion of them can be used as seeds for text generation.
Therefore, most of the sentences can be used as real seeds:
\begin{equation}
    \small
    \begin{gathered}
        H_{real}=\sigma\big(W_{emb}(S_{1}) + W_{pos}(S_{1})\big)\\
        \approx\sigma\big(W_{emb}(S_{n}) + W_{pos}(S_{n})\big) \in \mathbb{R}^{L\times d}
    \end{gathered}
    \label{equation4}
\end{equation}
where $L$ and $d$ represent sequence length and embedding dimensions.
$H_{real}$ from the real text can be viewed as continuous spaces, similar to images, and the well-pretrained seed interpretation model can predict the next sentence $S_{n+1}$ properly as illustrated in Figure~\ref{seedInterpretationModel} (right):
\begin{equation}
  \small
  S_{n+1}=f_{\theta}\Big(\sigma\big(W_{emb}(S_{n}) + W_{pos}(S_{n})\big)\Big) = f_{\theta}(H_{real}) \in \mathbb{Z}^{L}
  \label{equation5}
\end{equation}
As a result, text synthesis is carried out as the seed passes through the seed interpretation model to predict the subsequent sentence.

\subsubsection*{Applying to Unconditional Text Synthesis}
\begin{figure}[t]
  \centering
  \includegraphics[width=1\columnwidth]{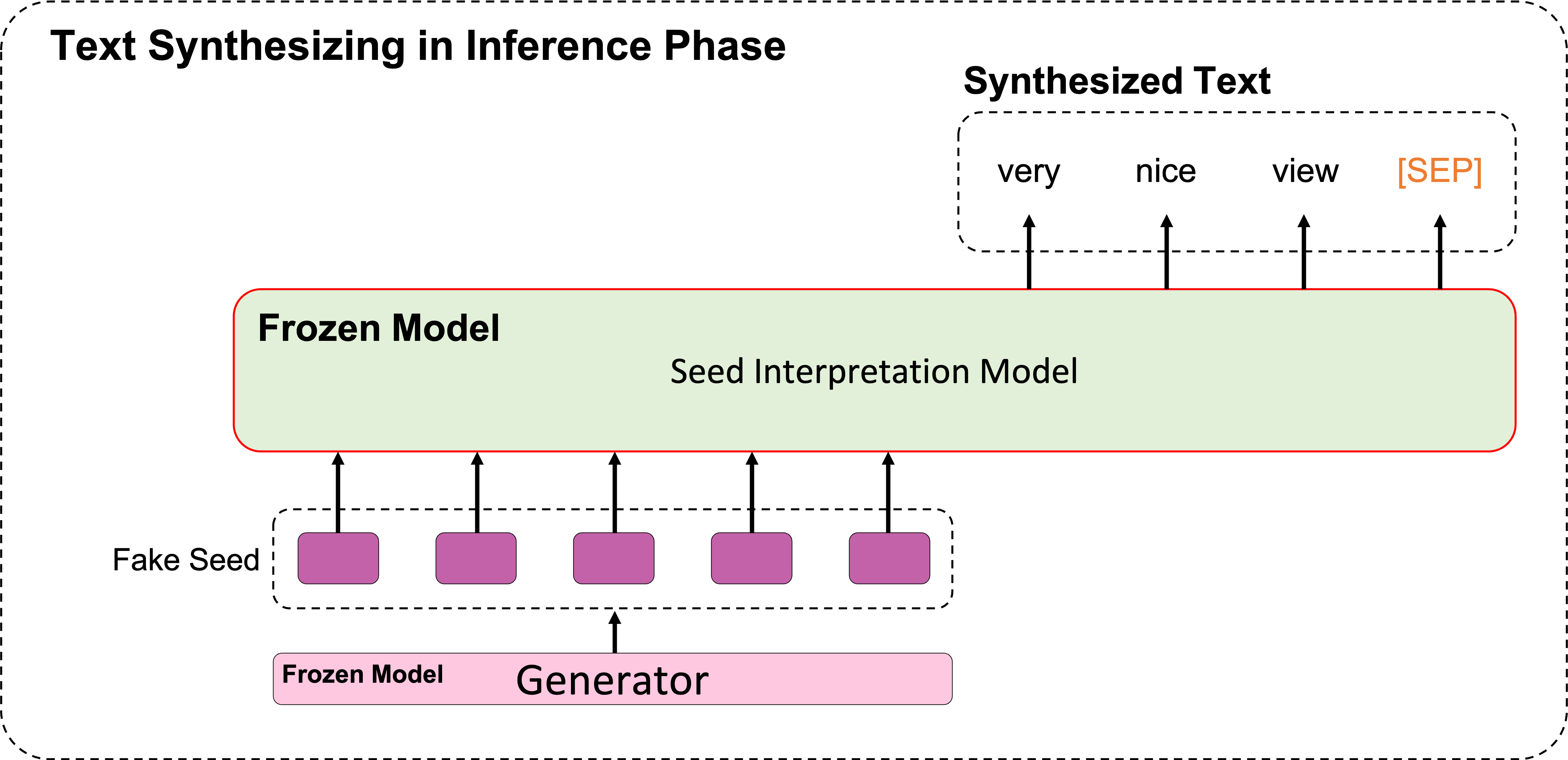}
  \caption{Illustration of text synthesizing method using the seed interpretation model in the inference phase.}
  \label{seedInterpretationModelInference}
\end{figure}
Here, we assume that the training of the TESGAN framework, including adversarial training, is fully completed and describe how the seed interpretation model synthesizes text during the inference phase.
Let $f_{\theta}^{*}(\cdot)$ and $g_{\phi}^{*}(\cdot)$ denote the frozen seed interpretation model and frozen generator, respectively.
We can now synthesize text during the inference stage using the well-trained $f_{\theta}^{*}(\cdot)$ and $g_{\phi}^{*}(\cdot)$.
At this point, $g_{\phi}^{*}(\cdot)$ will generate a fake seed $H_{fake}$ with the same dimensions as $H_{real}$, as shown in Equation~\ref{equation6}.
Then, $f_{\theta}^{*}(\cdot)$ can synthesize text using the fake seed, as shown in Figure~\ref{seedInterpretationModelInference}.
In other words, if the generator can skillfully create fake seeds $H_{fake}$ that imitate the distributions of $H_{real}$, then $H_{fake}$ can also generate appropriate subsequent sentences (a.k.a synthetic text).
However, no matter how excellently $H_{fake}$ is generated by the generator, it is useless if it cannot be interpreted; therefore, training the seed interpretation model is crucial. 
We use the pre-trained GPT-2~\cite{noauthororeditor}\footnote{\url{https://huggingface.co/docs/transformers/model\_doc/gpt2}\label{gpt2}} model and fine-tune it with multi-turn text data to serve as the seed interpretation model.
In addition, this model is only used to provide $H_{real}$ from the real text with frozen parameters during adversarial training.
Thus, the seed interpretation model never affects the training of the generator and the discriminator during adversarial training, and vice versa.
More detailed specifications of the seed interpretation model are explained in Appendix~\ref{appendixA}.

Synthesizing text based on the generated fake seeds $H_{fake}$ by the generator is entirely different from autoregressive prompting.
This is because prompting methods~\cite{DBLP:journals/corr/abs-2109-01652,ouyang2022training,chung2022scaling} function by providing discrete tokens as input to a model that generates the next tokens based on the previous one.
On the other hand, the generator of TESGAN creates continuous spaces $H_{fake}$ for synthesizing text from random noise, enabling unconditional text synthesis without explicit human instruction.
Furthermore, research is actively being conducted to leverage continuous spaces (learnable query) for flexible model training~\cite{lester-etal-2021-power,alayrac2022flamingo,li2023blip2,dai2023instructblip}.
Most research explores various methods, including using a fixed learned query after model training or memorizing multiple learned queries and selecting them selectively as needed in different situations.
However, the TESGAN framework differs from the mentioned studies in that its primary objective is to generate appropriate queries through the gernerator to produce appropriate sentences.

\subsection{Generator}
\begin{figure}[tb!]
  \centering
  \includegraphics[width=1\columnwidth]{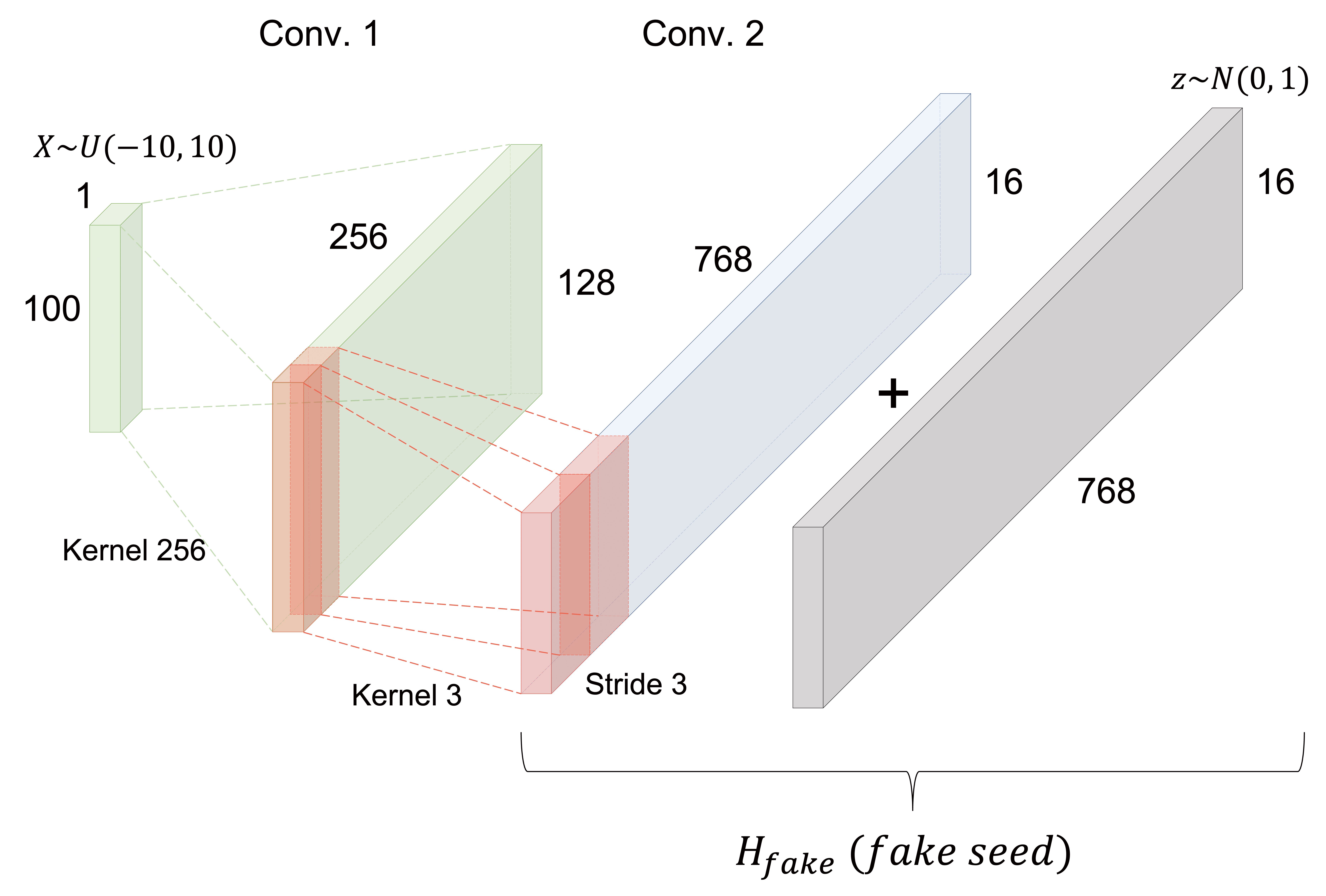}
  \caption{Illustration of the generator. P-TESGAN makes perturbed seeds by adding zero-centered normal distribution noise $z$ (gray) to the output (blue) from the generator.}
  \label{generator}
\end{figure}
Both real and fake seeds ($H_{real}$ and $H_{fake}$) are essential for adversarial training.
Real seeds can be obtained from the real text, as described in the seed interpretation model, and fake seeds are generated by the generator. 
In most text-GAN models reported so far, fake sentences are obtained from a text-based pre-trained autoregressive generator.
Consequently, data memorization occurs, where several synthetic sentences reproduce the training data.
To prevent data memorization, our generator does not use a pre-trained autoregressive-based model and does not explicitly reference the text in the training data during adversarial training. 
Our generator aims to create suitable fake text embedding spaces $H_{fake}$ in an unsupervised manner (GAN framework) by referencing real text continuous spaces $H_{real}$.

As shown in Figure~\ref{generator}, the generator $g_{\phi}(\cdot)$ is composed of two convolutional layers and generates seeds from the uniform distribution noise $X$ within an interval of $[-10,10)$ to create diverse forms of the seeds.
Additionally, using random noise has the advantage of not having to select the first token in the text synthesis process after model learning.
The final $H_{fake}$ can be obtained by Equation~\ref{equation6}:
\begin{equation}
    \small
    H_{fake}=g_{\phi}\big(X\sim U(-10,10)\big) \in \mathbb{R}^{L\times d}
    \label{equation6}
\end{equation}
where $L$ and $d$ represent sequence length and embedding dimensions.
As a result, the embedding space created by the generator has the same dimension as the real seed.
Moreover, we also compare an additional model, perturbed TESGAN (P-TESGAN).
P-TESGAN creates perturbed seeds by adding zero-centered normal distribution noise $z$ to the generator output.
P-TESGAN is expected to learn more robustly by perturbating the generator output.
Please refer to Appendix~\ref{appendixA} for detailed model information.

\subsection{Objective Functions}
\label{objectiveFunctions}
Since the generator does not refer to text during adversarial training, its performance is determined by the loss of the discriminator.
Thus, we propose four types of loss to update the parameters of the generator and the discriminator.

\subsubsection{Discriminators}
\label{discriminatorSection}
\begin{figure*}[tb!]
  \begin{subfigure}{0.5\textwidth}
    \centering
    \includegraphics[width=1\linewidth]{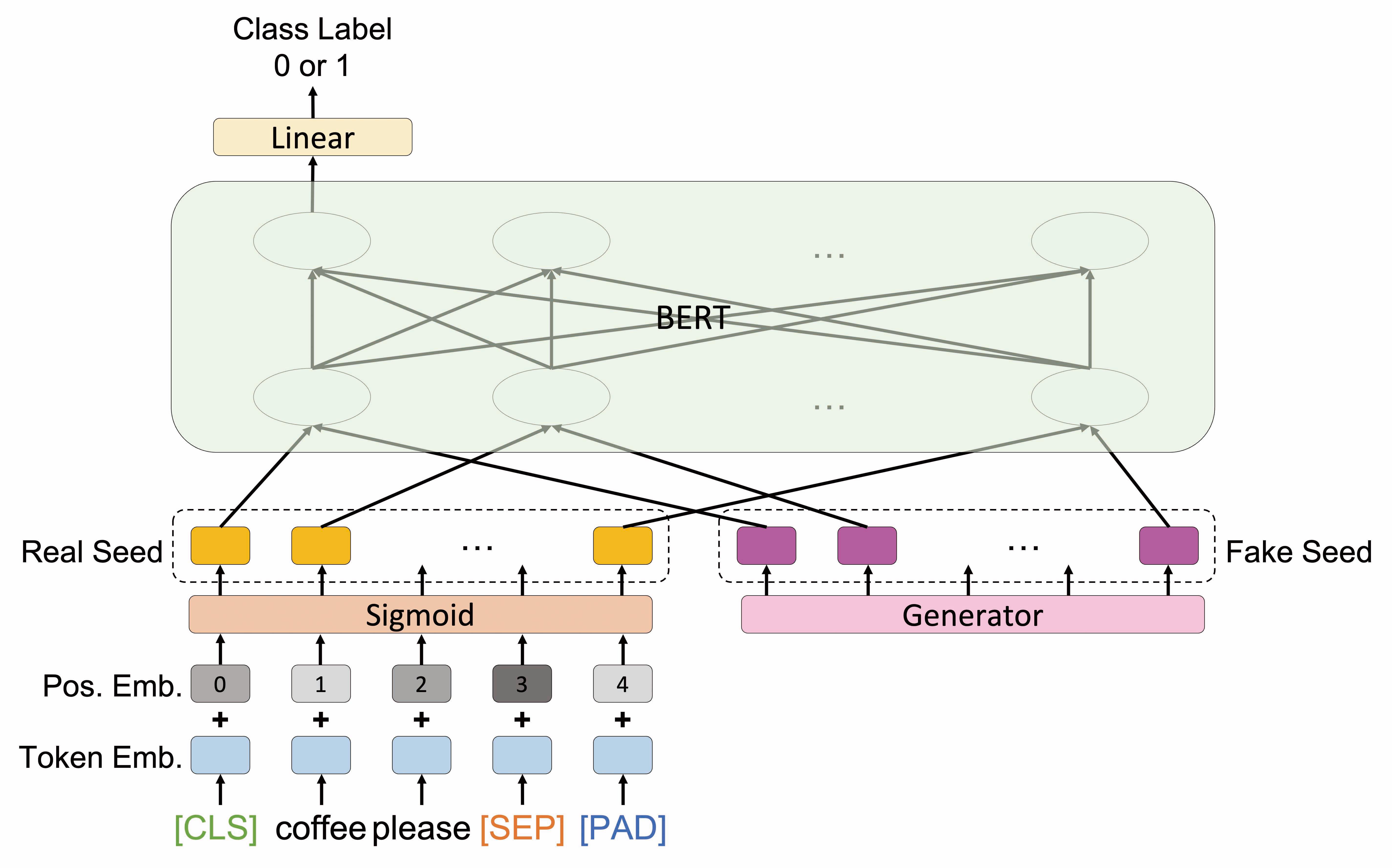}
    \caption{Seed Structure Discriminator (SSD)}
    \label{ssd}
  \end{subfigure}
  \begin{subfigure}{0.5\textwidth}
    \centering
    \includegraphics[width=1\linewidth]{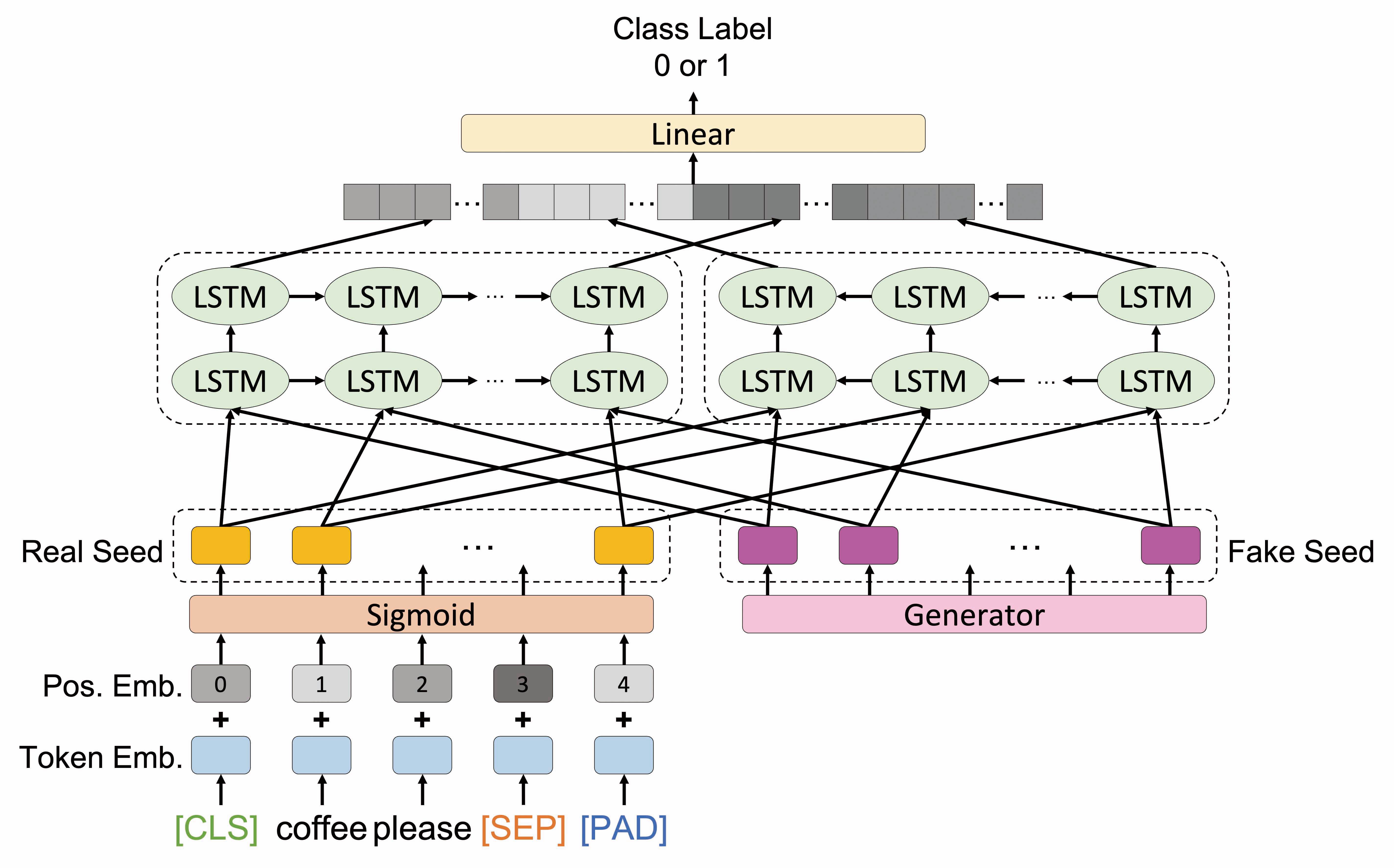}
    \caption{Seed Order Discriminator (SOD)}
    \label{sod}
  \end{subfigure}
  \caption{Illustrations of the two discriminators.
           SSD predicts whether the seed is real or fake using the $[CLS]$ special token's feature.
           SOD considers both forward and backward contexts of the seed.}
  \label{discriminators}
\end{figure*}
Sentence structure is important for constructing a complete sentence.
Since the real seeds are made from perfect sentences, they maintain structural representations of sentences.
Therefore, the fake seeds should capture the structural features of the real seeds.
As shown in Figure~\ref{ssd}, we use Bidirectional Encoder Representations from Transformer (BERT) \cite{devlin-etal-2019-bert} $d_{\alpha}(\cdot)$ called Seed Structure Discriminator (SSD) to capture the structural features of sentences, and the first hidden state is used to predict whether the seed is real or fake.

The order of tokens is also important for constructing sentences.
It is possible to predict whether a sentence's order representation of a seed is correct because both real and fake seeds have a dimension of (\textit{\textbf{sequence length} * embedding dimensions}).
To do this, as shown in Figure~\ref{sod}, we use Bidirectional LSTM $d_{\beta}(\cdot)$ called Seed Order Discriminator (SOD) to consider both forward and backward directions of sentences.
The concatenation of the first and the last hidden states is used to predict whether the seed is real or fake.

During adversarial training, both discriminators are trained to predict whether the seeds are real (label 1) or fake (label 0), while the generator is trained to fool the discriminators by predicting fake seeds as 1.
The loss function of the discriminators is defined by the following equation, which updates both the discriminators and the generator:
\begin{equation}
  \small
  \begin{gathered}
    \mathcal{L}_{D} = -\frac{1}{N}\sum_{i=1}^{N}\big[y_i\log{x_i} + (1-y_i) \log{(1 - x_i)} \big] \\
    x=predicted, y=target
  \end{gathered}
  \label{discriminator}
\end{equation}
Additional information regarding the size and descriptions of the discriminators can be found in Appendix~\ref{appendixA}.

\subsubsection{Generator Helpers}
\begin{figure}[tb!]
  \centering
  \includegraphics[width=1\columnwidth]{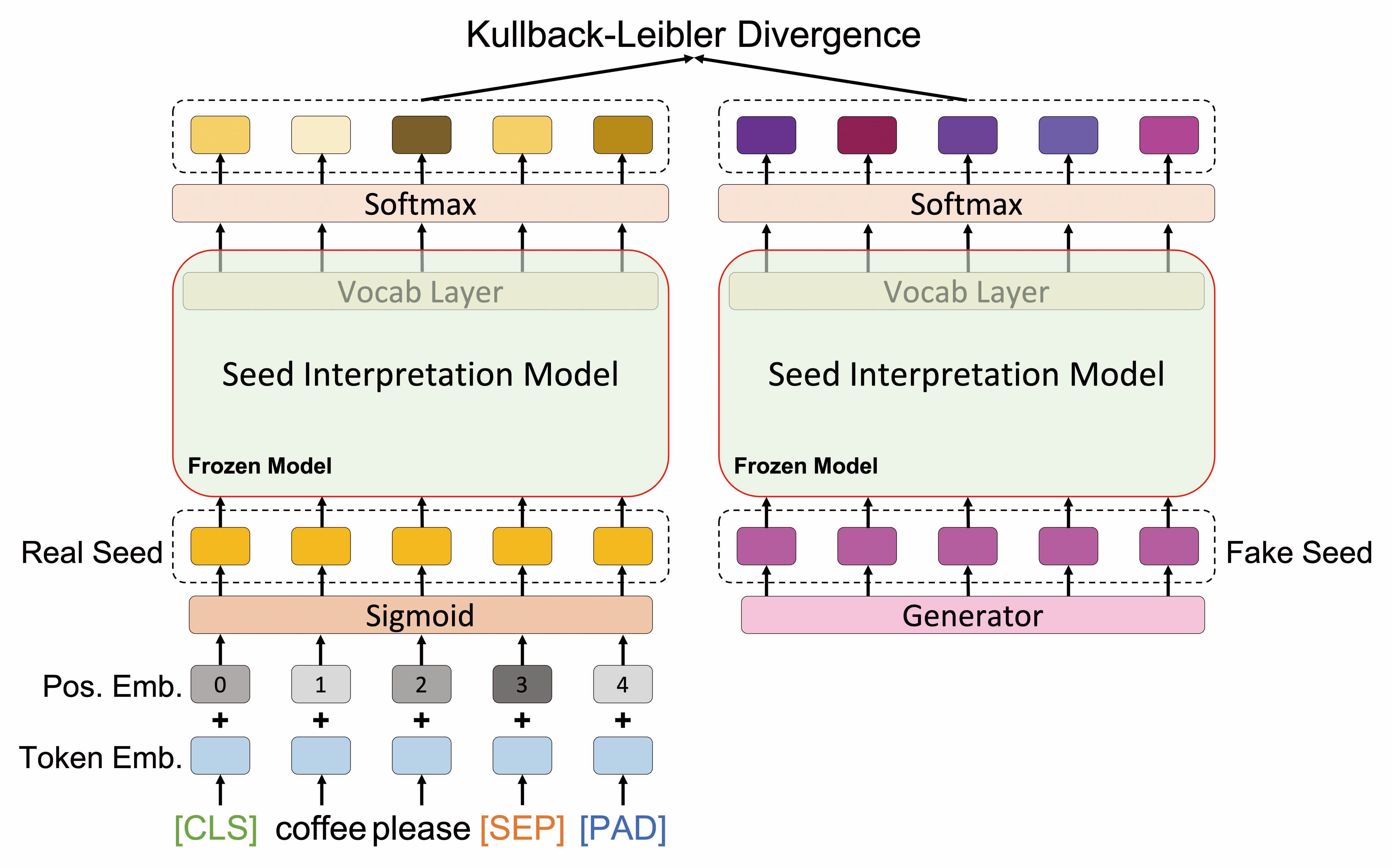}
  \caption{Illustration of Seed Distribution Prediction (SDP).
           SDP is used to enhance the fake seeds of the generator during adversarial training by minimizing the distance between real and fake seed distributions.}
  \label{sdp_img}
\end{figure}
During adversarial training, it is challenging for the generator to learn solely from the discriminators introduced in Section~\ref{discriminatorSection}.
Therefore, in this section, two auxiliary tasks are introduced to aid the training of the generator.

Capturing the distribution of the text embedding space is important, and for this purpose, we employ Seed Distribution Prediction (SDP).
However, since the text embedding space cannot be directly used as a probability distribution, the output of the seed interpretation model is utilized.
Specifically, when a seed passes through the frozen seed interpretation model, the output dimension of (\textit{sequence length * vocabulary size}) is obtained through the softmax function, which can be used as a probability distribution.
The loss is calculated using the Kullback-Leibler divergence between the distributions of the real and the fake seeds.
SDP is used solely for updating the generator during adversarial training:
\begin{equation}
  \small
  \mathcal{L}_{SDP} = \sigma\big(f_{\theta}^{*}(H_{real})\big)\log{\frac{\sigma\big(f_{\theta}^{*}(H_{real})\big)}{\sigma\big(f_{\theta}^{*}(H_{fake})\big)}}
  \label{sdp}
\end{equation}
where the $\sigma$ and $f_{\theta}^{*}(\cdot)$ mean \textit{softmax} function and the frozen seed interpretation model.
More detailed figure of SDP is illustrated in Figure~\ref{sdp_img}.

The sentences used as the seeds are composed of tokens explained in Equation~\ref{equation2}.
Additionally, we apply Seed Frame Prediction (SFP) since the structures of seeds are somewhat formalized.
Therefore, we calculate the Mean Absolute Error (MAE) and Mean Squared Error (MSE) to make the form of a fake seed similar to a real one.
If we train the fake seeds using MAE and MSE, the fake seeds from the generator can become blurred.
However, the loss of SFP is relatively small compared to that of SSD, SOD, and SDP; therefore, SFP does not adversely affect the generator.
SFP is used only for updating the generator during adversarial training:
\begin{equation}
  \small
  \begin{gathered}
    \mathcal{L}_{SFP} = \|\mu_{r} - \mu_{f}\|_2^2 + \|H_{real} - H_{fake}\|_1\\
    \mu_r=avg(H_{real}), \mu_f=avg(H_{fake})
  \end{gathered}
  \label{sfp}
\end{equation}
The full loss function, including the seed interpretation model's loss, is described in Appendix~\ref{losses}.

\section{Text Synthesis Experiments}
\subsection{Dataset}
In this experiment, we use two datasets consisting of multi-turn sentences to train the seed interpretation model and perform the text synthesis experiment.

\noindent\textbf{DailyDialog}\footnote{\url{http://yanran.li/dailydialog}}~\cite{li-etal-2017-dailydialog} is multi-turn conversation data used for training open-domain dialogue generation models.
It consists of chit-chat-style multi-turn English conversations, and we select this data for domain-independent text synthesis.
This dataset is used to evaluate the performance of TESGAN and other baselines.

\noindent\textbf{IMDb}\footnote{\url{https://huggingface.co/datasets/imdb}}~\cite{maas-EtAl:2011:ACL-HLT2011} contains highly polar movie reviews and is widely used for sentiment classification tasks.
Each human-written movie review consists of several sentences, and we used this data as multi-turn data.
This dataset is rougher and has a larger volume compared to DailyDialog.
We evaluate the general applicability by synthesizing sentences based on IMDb-trained TESGAN.
Statistics of the two datasets are shown in Appendix~\ref{datasetStats}.

\subsection{Training Steps}
TESGAN training has two steps.
\begin{algorithm}[t]
  \caption{Text Embedding Space GAN}
  \label{algorithm1}
  \textbf{Require}: Seed interpretation model $f_{\theta}$; Generator $g_{\phi}$; BERT discriminator $d_{\alpha}$; LSTM discriminator $d_{\beta}$; Multi-turn data $D=\{S_{1:N}\}$; Sentence data $S_{i}=\{w^{i}_{1:L}\}$.

  \begin{algorithmic}[1]
      \STATE Pre-train $f_{\theta}$ using $D$.
      \STATE Initialize $g_{\phi}$, $d_{\alpha}$, $d_{\beta}$ with random weights\\ $\phi, \alpha, \beta \sim N(0, 0.08)$.
      \STATE Freeze the $f_{\theta}$.
      \WHILE{TESGAN converges}
      \FOR{d-steps (during odd epoch)}
      \STATE Get real data from $f_{\theta}$ using $S$ with positive label 1.
      \STATE Make fake data from $g_{\phi}$ with negative label 0.
      \STATE Update $\alpha$ and $\beta$ via results of $d_{\alpha}$ and $d_{\beta}$.
      \ENDFOR
      \FOR{g-steps}
      \STATE Make fake data from $g_{\phi}$ with positive label 1.
      \STATE Calculate SDP and SFP.
      \STATE Update $\phi$ via results of $d_{\alpha}$, $d_{\beta}$, SDP and SFP.
      \ENDFOR
      \ENDWHILE
  \end{algorithmic}
\end{algorithm}
First, the seed interpretation model must be pre-trained with multi-turn data to interpret the seeds.
In the performance and general applicability experiments, we train the model on the 11k and 25k multi-turn sets of DailyDialog and IMDb, respectively, as shown in Table~\ref{table1}. 
Then, the model that achieves the highest BLEU-4 score in the validation set is selected in each experiment.

The second step is adversarial training.
After pre-training the seed interpretation model, the generator and the discriminator learn through adversarial training.
For adversarial training, real and fake seeds are created by the embedding part of the frozen seed interpretation model and the generator, respectively.
Since real seeds can be generated from a significant number of sentences, all 87k and 300k sentences in each training set used in the experiment mentioned above are used to create the real seeds via Equation~\ref{equation2}. 
We also generate the same number of fake seeds as real seeds for adversarial training, and the following equation represents what the discriminator and generator aim to optimize during adversarial training:
\begin{equation}
  \small
  \begin{gathered}
    \mathcal{D}_{\mathcal{L}} = \max_{\alpha, \beta}{\mathbb{E}_{x \sim H_{real}}\Big[\log{d_{\alpha,\beta}(x)}\Big]} \\
    \mathcal{G}_{\mathcal{L}} = \max_{\phi}{\mathbb{E}_{z}\Big[\log{d_{\alpha,\beta}(g_{\phi}(z))}\Big]} + \mathcal{L}_{SDP} + \mathcal{L}_{SFP}
  \end{gathered}
  \label{loss}
\end{equation}
where $d_{\alpha, \beta}$ means SSD, SOD respectively.
$\mathcal{D}_{\mathcal{L}}$ implies updating the parameters of the discriminator to accurately predict real seeds as 1 from the perspective of real seeds.
$\mathcal{G}_{\mathcal{L}}$ also means updating the generator so that the discriminator predicts the fake seeds created by the generator as 1.
This approach helps partially resolve the learning imbalance problem between the generator and the discriminator~\cite{NIPS2014_5ca3e9b1}.
Further discussion of the above pseudocode and optimization methods is covered in Section~\ref{generatorAndTrainingStrategy}.
Lastly, hyperparameters and experiment setup are described in Appendix~\ref{hyperparameters}.

\subsection{Evaluation Metric}
\label{metrics}
Target-oriented evaluation metrics, such as BLEU and ROUGE~\cite{lin-2004-rouge}, are not suitable for evaluating synthetic text.
This is because each synthesized sentence from random noise has no corresponding target, and the generative models aim to synthesize plausible data based on real data distribution without copying the training data.
Therefore, we employ several metrics that can evaluate unconditional text generation.

\subsubsection{Fréchet BERT Distance (FBD)}
\citet{dAutume2019TrainingLG} proposed Fréchet Embedding Distance (FED) to evaluate the quality of synthetic text, inspired by Fréchet Inception Distance (FID) \cite{NIPS2017_8a1d6947}.
\citet{alihosseini-etal-2019-jointly} proposed FBD, an improved version of FED, to measure the quality and diversity of synthesized text using a pre-trained BERT.
The features of real and synthesized text obtained by the pre-trained BERT are assumed to have Gaussian distributions, and FBD is the distance between them:
\begin{equation}
  \small
  FBD=\sqrt{\|\mu_{r} - \mu_{f}\|_{2}^{2}+tr\big(\Sigma_r + \Sigma_f - 2(\Sigma_r\Sigma_f)^{0.5}\big)}
  \label{equation7}
\end{equation}
where $\mu$ and $\Sigma$ show the mean vectors and the covariance matrices of the real and fake seed features.

\subsubsection{Multi-Sets-Jaccard (MSJ)}
Each synthesized sentence has no corresponding target;
thus, we select MSJ~\cite{alihosseini-etal-2019-jointly}, which calculates the score between real and synthesized text sets.
The Jaccard Index determines the similarity of two sets, calculating the ratio of the cardinality of their intersection to that of their union.
Inspired by the Jaccard Index, MSJ focuses on the similarity of the n-gram frequencies of text in the two sets, $s_r$ and $s_f$, which are the real and synthesized text sets, respectively:
\begin{equation}
  \small
  MSJ_n = \frac{\sum_{g\in G_n}\min\big(C_n(g,s_r),C_n(g,s_f)\big)}{\sum_{g\in G_n}\max\big(C_n(g,s_r),C_n(g,s_f)\big)}
  \label{equation8}
\end{equation}
where $G_n$ and $C_n(g, s)$ mean the n-gram in $s_r\cup s_f$ and the normalized counts of the n-gram in set $s$.
Additionally, this n-gram-based synthesized sentence evaluation method is a common approach in the field of unconditional text generation~\cite{Yu_Zhang_Wang_Yu_2017,DBLP:journals/corr/PressBBBW17,DBLP:conf/iclr/FedusGD18}.

\subsubsection{Language Model score (LM)}
\citet{dAutume2019TrainingLG,DBLP:conf/iclr/CacciaCFLPC20} proposed LM, which can evaluate the quality of generated samples using a well-trained language model.
LM measures the quality of generated samples, meaning that scores of the bad samples are poor under a well-trained language model.
We select the pre-trained GPT-2\footref{gpt2} as a well-trained language model.
LM is calculated as the cross-entropy results between the output and input of GPT-2.

\subsubsection{Data Synthesis Ratio (DSR)}
DSR considers not only the data memorization ratio between the training and synthesized data but also synthetic diversity itself.
Short sentences identical to training data, such as "\textit{I'm fine}", can be synthesized by coincidence.
Therefore, sentences longer than two-thirds of the maximum sentence length that perfectly reproduce the training data are considered memorized data.
Considering these conditions, we can calculate DSR using the following equation:
\begin{equation}
  \small
  \begin{gathered}
      R_{syn} = \frac{|S_{syn} - S_{train}|}{|S_{syn}|}, R_{unq} = \frac{|S_{unq}|}{|S_{syn}|} \\
      DSR = \frac{2*R_{syn}*R_{unq}}{R_{syn}+R_{unq}}\\
  \end{gathered}
  \label{equation9}
\end{equation}
where $S_{syn}$ and $S_{train}$ indicate synthesized and training text set respectively.
$S_{unq}$ means the set of unique sentences of synthesized text results.
If the synthesized sentences in $S_{syn}$ do not reproduce any of the sentences in $S_{train}$, $R_{syn}$ would be 1.
Similarly, if the synthesized text in $S_{syn}$ is all unique, $R_{unq}$ will be 1.
The final DSR is calculated as the harmonic mean of $R_{syn}$ and $R_{unq}$ ratios.

\subsubsection{Self-BLEU (SBL)}
\citet{10.1145/3209978.3210080} first proposed SBL to measure diversity of token combination.
The original BLEU evaluates the degree of n-gram overlap (similarity) between one hypothesis sentence and multiple reference sentences.
However, unconditionally generated text does not have specific targets, so it is not suitable for BLEU evaluation.
SBL is widely used to solve this problem.
SBL can evaluate n-gram-level similarity by regarding one sentence as a hypothesis and the rest as references in a synthetic text set.
Since SBL evaluates based on the generated text set itself, it is not able to evaluate the quality of the synthetic text, but it is possible to evaluate the diversity of token combinations based on n-gram.
Additionally, the difference between SBL and DSR lies in their evaluation criteria.
DSR assesses data memorization by comparing the generated text set with the training dataset, while also considering the diversity of not n-gram-based but generated complete sentences themselves.

\subsection{Baselines}
\begin{figure*}[htb!]
  \begin{subfigure}{.33\textwidth}
    \centering
    \includegraphics[width=1\linewidth]{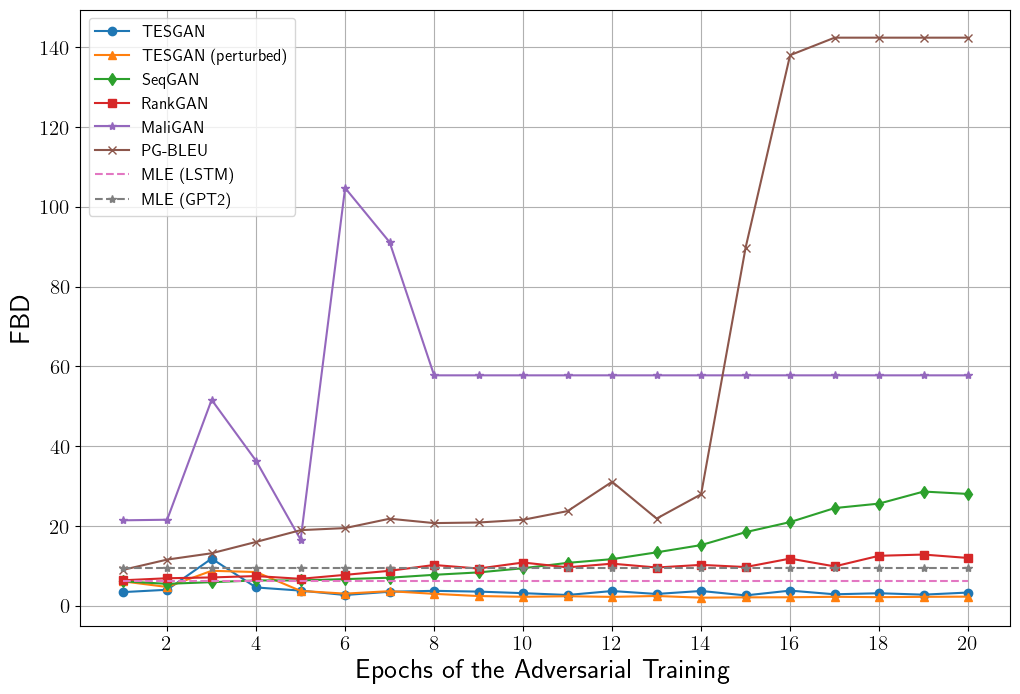}
    \caption{FBD $\downarrow$}
    \label{figure3a}
  \end{subfigure}
  \begin{subfigure}{.33\textwidth}
    \centering
    \includegraphics[width=1\linewidth]{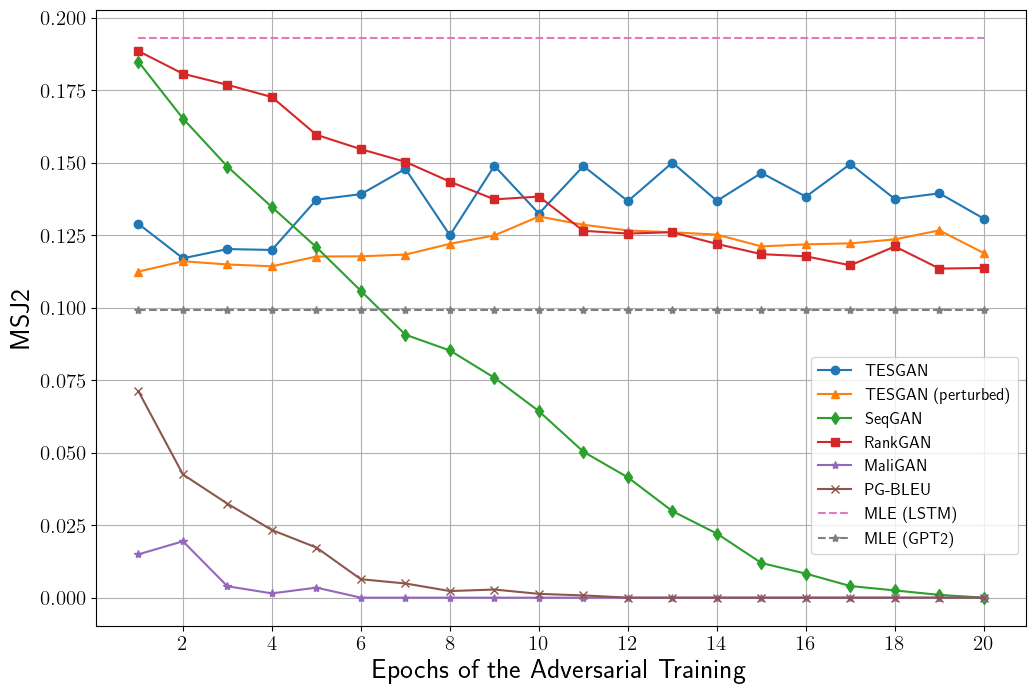}
    \caption{MS-Jaccard2 $\uparrow$}
    \label{figureMSJ2}
  \end{subfigure}
  \begin{subfigure}{.33\textwidth}
      \centering
      \includegraphics[width=1\linewidth]{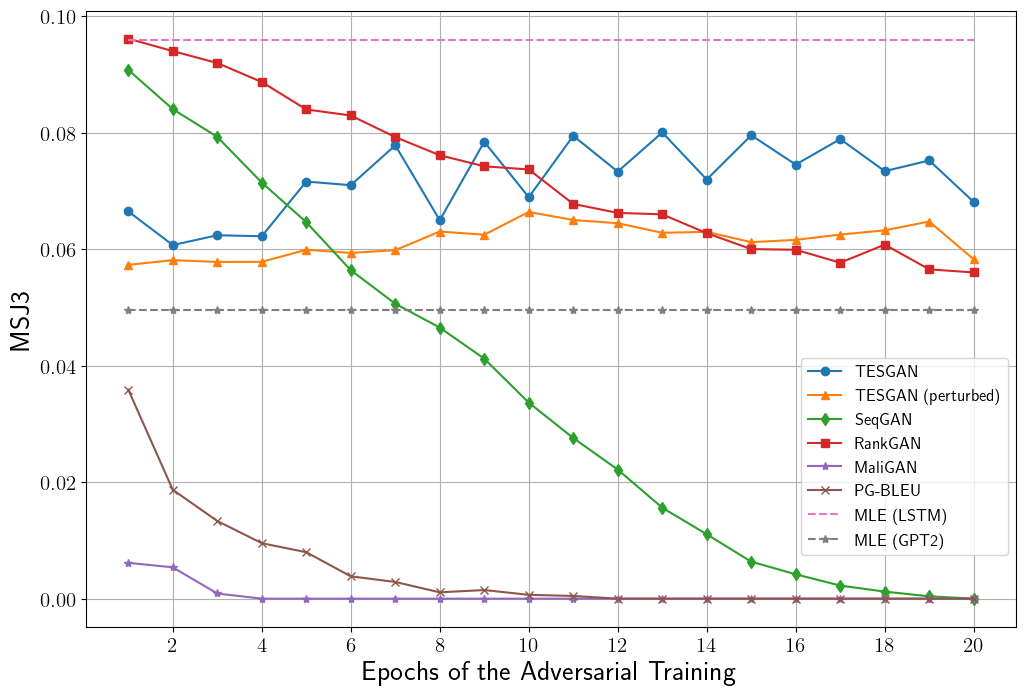}
      \caption{MS-Jaccard3 $\uparrow$}
      \label{figureMSJ3}
  \end{subfigure}
  \begin{subfigure}{.33\textwidth}
    \centering
    \includegraphics[width=1\linewidth]{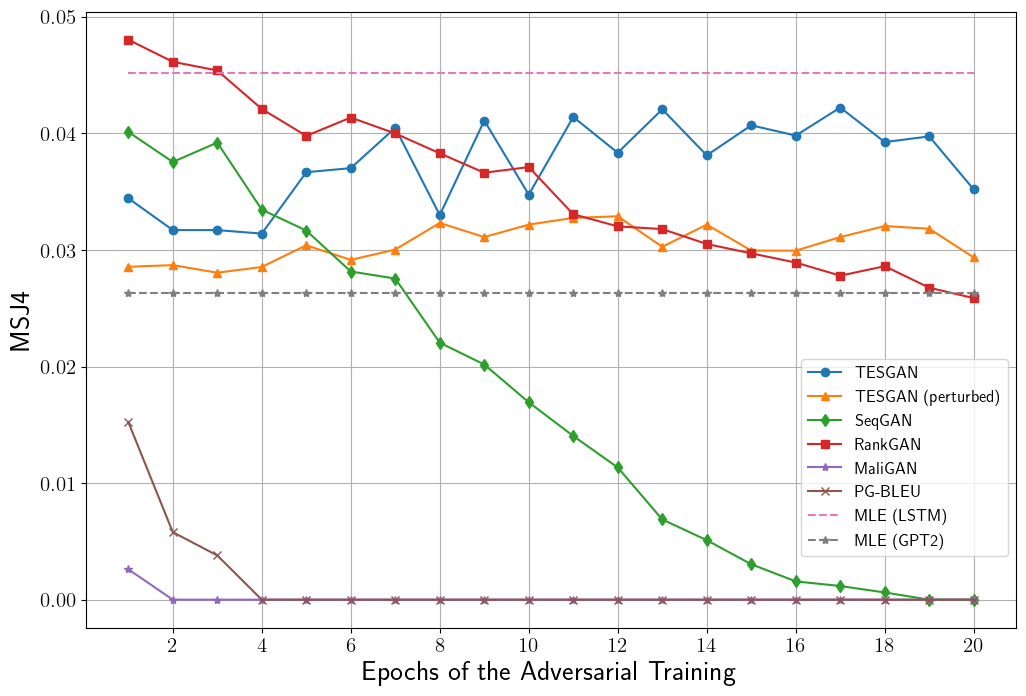}
    \caption{MS-Jaccard4 $\uparrow$}
    \label{figureMSJ4}
  \end{subfigure}
  \begin{subfigure}{.33\textwidth}
    \centering
    \includegraphics[width=1\linewidth]{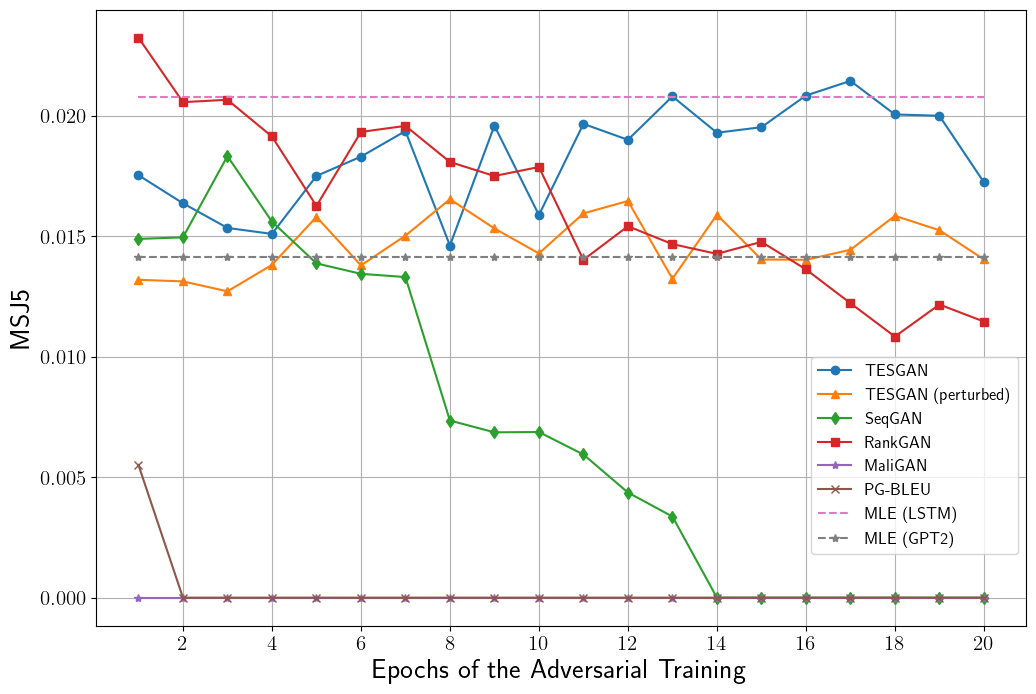}
    \caption{MS-Jaccard5 $\uparrow$}
    \label{figure3e}
  \end{subfigure}
  \begin{subfigure}{.33\textwidth}
      \centering
      \includegraphics[width=1\linewidth]{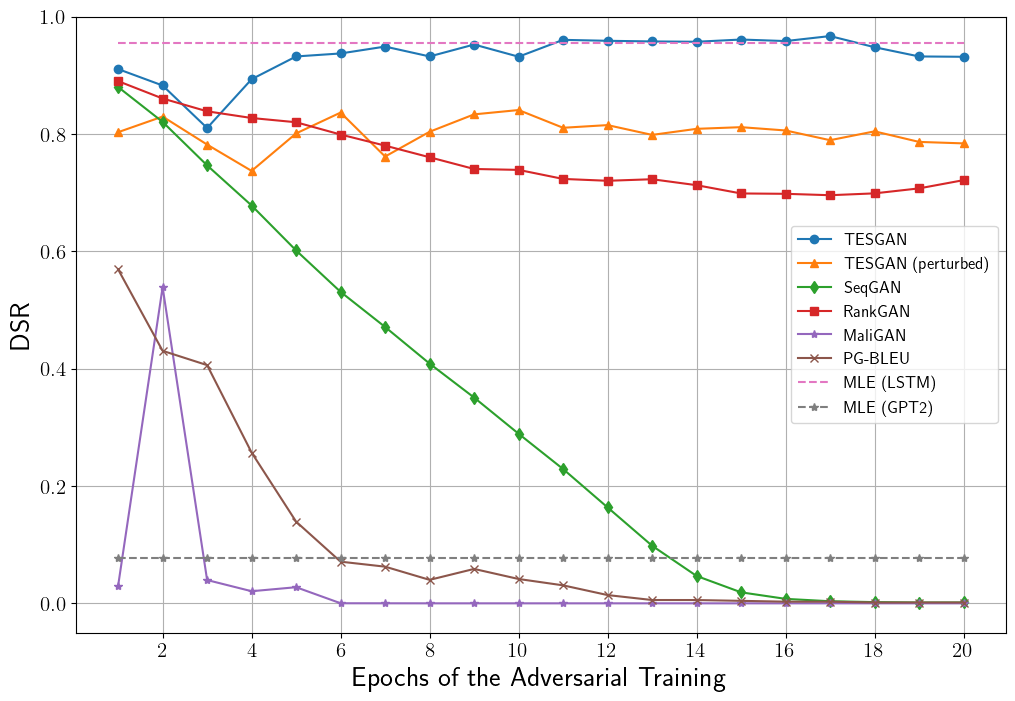}
      \caption{DSR $\uparrow$}
      \label{figure3f}
  \end{subfigure}
  \caption{Illustration showing the results of the text-GAN models.
  In previous research, adversarial training is conducted after the generator pre-training. 
  MLE is represented as a baseline because it is a supervised pre-trained generator without adversarial training.}
  \label{figure3}
\end{figure*}

In this paper, we compare our two models with the following approaches: LSTM-based Maximum Likelihood Estimation (MLE-L), PG-BLEU, SeqGAN, RankGAN, and MaliGAN.
MLE-L represents the pre-training result of the generator, which all the other models undergo before adversarial training. 
The pre-trained generators with the lowest loss in the validation set were chosen for each method, including MLE-L. 
We compare these models with our original TESGAN and P-TESGAN, which is trained by adding zero-centered normal distribution noise $z$ to the generator's output.
We also evaluate the GPT-2-based pre-trained seed interpretation model (MLE-G) used in the TESGAN framework.
Since MLE-based models are trained without adversarial training, they are shown as baselines in Figure~\ref{figure3}.
Finally, to demonstrate that the results of the TESGAN-based models are not solely dependent on the seed interpretation model but rather on seeds created by the generator, we present the outcomes when using Gaussian random noise as input for the seed interpretation model.

\section{Results}
\begin{table*}[htb!]
  \centering
  \small
  \begin{tabular}{lrrrrrrr}
      \hline
      Method & FBD\textsuperscript{Q,D} $\downarrow$ & MSJ4\textsuperscript{Q,D} $\uparrow$ & MSJ5\textsuperscript{Q,D} $\uparrow$ & DSR\textsuperscript{M} ($R_{syn}, R_{unq}$) $\uparrow$ & LM$\ast$\textsuperscript{Q} $\downarrow$ & SBL3$\ast$\textsuperscript{D}$\downarrow$  & SBL4$\ast$\textsuperscript{D} $\downarrow$ \\
      \hline\hline
      TESGAN (ours) & 2.899 & 0.042 & 0.021 & \textbf{0.967} (\textbf{1}, 0.936) & 4.236 & 0.743 & 0.623 \\
      P-TESGAN (ours) & \textbf{2.274} & 0.032 & 0.014 & 0.841 (0.997, 0.727) & 3.642 & 0.790 & 0.702 \\
      SeqGAN~\cite{Yu_Zhang_Wang_Yu_2017} & 6.153 & 0.040 & 0.015 & 0.880 (0.883, 0.877) & 5.094 & 0.420 & 0.266 \\
      RankGAN~\cite{10.5555/3294996.3295075} & 6.409 & \textbf{0.048} & \textbf{0.023} & 0.890 (0.895, 0.886) & 5.123 & 0.446 & 0.290 \\
      MaliGAN~\cite{https://doi.org/10.48550/arxiv.1702.07983} & 21.436 & 0.003 & 0 & 0.030 (\textbf{1}, 0.015) & - & - & - \\
      PG-BLEU~\cite{Yu_Zhang_Wang_Yu_2017} & 9.002 & 0.015 & 0.006 & 0.569 (0.555, 0.584) & 4.584 & 0.628 & 0.484 \\
      \hline
      MLE-L~\cite{Yu_Zhang_Wang_Yu_2017} & 6.284 & 0.045 & 0.021 & 0.955 (0.925, \textbf{0.987}) & 5.168 & \textbf{0.403} & \textbf{0.242} \\
      MLE-G & 9.592 & 0.026 & 0.014 & 0.078 (\textbf{1}, 0.040) & \textbf{3.543} & 0.948 & 0.944 \\
      \hline
      Random Noise $\dagger$ & 14.142 & 0.016 & 0.006 & 0.930 (1, 0.869) & 4.562 & 0.516 & 0.404 \\
      \hline
  \end{tabular}
  \caption{Performance of the models. P-TESGAN denotes the perturbed TESGAN.
        $\dagger$ is the result of directly entering Gaussian random noise as an input to the seed interpretation model. 
        The second group of models consists of autoregressive models.
        $\ast$ denotes a metric not considered when selecting the best model.
        - denotes that the confidence of the result is low because the quality of the synthesized sentence is poor.
        The superscript of each metric represents what each metric can measure (Q: quality, D: diversity, M: data memorization).}
  \label{table2}
\end{table*}

\begin{table*}[t]
    \centering
    \small
    \begin{tabular}{|l|l|l|}
        \hline
        \textbf{TESGAN (17-epoch, DailyDialog)} & \textbf{P-TESGAN (10-epoch, DailyDialog)} & \textbf{Random Noise} \\
        \hline
        I'm so glad you finally got on the train. & Hello, Mr. Smith. I'm Mary. & Anything I have called three weeks\\
        I just lost my job. & I just want to tell you the truth. & Is Is Is Is Is Is Is\\
        Yeah. You mean the network connection? & It's the end of the world. & Left and go to go to go to go\\
        What happened? & What do you want to do in this company? & Mr Moon, Mr Moon\ldots Mr Moon\ldots\\
        So you have to wait for a while. & He just broke up with Ann. & are you have finished 6 items?\\
        \hline\hline
        \multicolumn{3}{|l|}{\textbf{TESGAN (18-epoch, IMDb)}} \\
        \hline
        \multicolumn{3}{|l|}{This is probably one of the best of the best of the series.} \\
        \multicolumn{3}{|l|}{I was bored to think about how stupid this movie was.} \\
        \multicolumn{3}{|l|}{"The Deadly Loved One" is the story of a rebellious college basketball} \\
        \multicolumn{3}{|l|}{I have to say, this is the worst film I have ever seen.} \\
        \multicolumn{3}{|l|}{I was very excited to see it, anticipating Christmas eve.} \\
        \multicolumn{3}{|l|}{This movie was one of the best of the year for me.} \\
        \hline
    \end{tabular}
    \caption{Example of unconditionally synthesized sentences. P-TESGAN denotes the perturbed TESGAN.}
    \label{table3}
\end{table*}

\subsection{Metric-based Evaluation}
\label{metricResults}
In this section, we compare the results of the synthesized text at every epoch of adversarial training using the metrics mentioned in Section~\ref{metrics}. 
This experiment was performed with models trained on the DailyDialog dataset.
Since Fréchet BERT Distance (FBD) and Multi-Sets-Jaccard (MSJ) require a real text corpus, the test set is used as the real text corpus.
Data Synthesis Ratio (DSR) is calculated with the training set as the data memorization ratio needs to be computed.

The FBDs of the TESGAN-based models are lower than the MLE-based autoregressive results, while the baselines increase after having the lowest value at the end of the first epoch of adversarial training, as shown in Figure~\ref{figure3a}.
In terms of MSJ, as shown in Figure~\ref{figureMSJ2}-\ref{figure3e}, the previous studies report lower values than the TESGAN-based models at the end of adversarial training, despite having higher results in the beginning.
On the other hand, MSJ results of the TESGAN-based models slightly increase during adversarial training.
Moreover, some MSJ5 results of the original TESGAN are higher than MLE-L during adversarial training, as shown in Figure~\ref{figure3e}.
In the case of DSR, as shown in Figure~\ref{figure3f}, the original TESGAN also increases during adversarial training, and some results are higher than MLE-L. 
On the other hand, the results of the previous studies decrease during adversarial training, resulting in lower values than the TESGAN-based models in the end.
As adversarial learning progresses, the results of the baselines deteriorate because the LSTM generator tends to generate only a few unique sentences.
Furthermore, we will explain the reasons why the MLE-G results of the GPT-2 base are relatively poor in the following section.

We chose the best model of each method considering the FBD, MSJ, and DSR results because these metrics evaluate quality, diversity, and data memorization.
We evaluated the text generated by each model per epoch using the metrics mentioned above and compared the best-performing models\footnote{Detailed results are shown in Appendix~\ref{baselinesPerformanceResults}}.
According to Table~\ref{epochResults} in Appendix~\ref{baselinesPerformanceResults}, we compare the baselines at 1-epoch with TESGAN and P-TESGAN at 17 and 10-epochs, respectively. 
After selecting the best models, we calculated the Language Model score (LM) and Self-BLEU (SBL) based on the text generated by each model.
As shown in Table~\ref{table2}, the TESGAN-based models show the highest results in FBD and DSR.
Also, the TESGAN-based models show comparable results in MSJ compared to the baselines and display the highest results among the adversarial-based methods in terms of LM score.
However, in terms of SBL, TESGAN-based models perform worse than the baselines.
In addition, the results of Gaussian random noise demonstrate that the TESGAN results are attributed to the seeds from the generator. 
The first group of Table~\ref{table3} shows the synthetic text by TESGAN and Gaussian random noise.
In conclusion, MLE-L is a supervised pre-trained generator applied before adversarial training, but most of the result curves of the prior methods showed lower performance than MLE-L during adversarial training.
On the other hand, our TESGAN-based models showed better results than MLE-L or improved performance during adversarial training. 
Finally, the results according to the epoch of the LM and SBL of each model are shown in Appendix~\ref{baselinesPerformanceResults}.

\subsection{Analysis of Autoregressive Models}
\label{armodelanalysis}
In this section, we will analyze the results of the MLE-based autoregressive models.
Other baseline models pretrain an LSTM-based generator before starting adversarial training, while the TESGAN framework employs a GPT-2-based pre-trained seed interpretation model. 
The results of the MLE-based models in Section~\ref{metricResults} are based on the evaluation of corpora generated in an autoregressive manner using the two pre-trained models.
MLE-based models generate sentences in an autoregressive manner, starting from a specific $[start\_token]$ and predicting the next token.
If the model predicts the next token in a greedy manner, all generated sentences would be identical, exhibiting deterministic behavior.
To prevent this, MLE-based models sample the next token based on the probability logits~\cite{Yu_Zhang_Wang_Yu_2017}.
This way, MLE-L results in diverse token choices since the logit probability differences are not large.
On the other hand, MLE-G training fits the data better than LSTM-based models, resulting in significantly larger differences in the logits of the next token.
As a consequence, MLE-G is relatively deterministic compared to MLE-L.
Therefore, when generating sentences without the use of the softmax temperature technique~\cite{44873}, MLE-G delivers high quality, but it struggles to produce a variety of sentences. 
In practice, sentences generated by MLE-G lack diversity, which led to relatively lower results in Section~\ref{metricResults}.
However, it is worth noting that while diversity may be lacking, the quality of the generated sentences is high, and this aspect will be demonstrated in the next section.

\subsection{Human Evaluation}
\begin{table}[tb!]
  \centering
  \begin{tabular}{lr}
      \hline
      Method & Avg. Score \\
      \hline
      TESGAN (ours) & \textbf{4.2} \\
      P-TESGAN (ours) & 3.4\\
      SeqGAN~\cite{Yu_Zhang_Wang_Yu_2017} & 2.4 \\
      RankGAN~\cite{10.5555/3294996.3295075} & 2.0 \\
      MaliGAN~\cite{https://doi.org/10.48550/arxiv.1702.07983} & 1.0 \\
      PG-BLEU~\cite{Yu_Zhang_Wang_Yu_2017} &  1.6 \\
      MLE-L~\cite{Yu_Zhang_Wang_Yu_2017} & 3.0 \\
      MLE-G & 3.8 \\
      \hline
  \end{tabular}
  \caption{Human evaluation scores (1 $\sim$ 5).}
  \label{humanEvaluation}
\end{table}
We conducted human evaluations based on the corpora generated by each model.
The corpora, comprising 50 randomly selected unique sentences that do not duplicate those from the training set, were assessed by 10 annotators.
We asked annotators to give higher scores to corpora that contained more natural sentences on a scale from 1 to 5.
The scores presented in Table~\ref{humanEvaluation} represent the average scores assessed by each person.
According to Table~\ref{humanEvaluation}, TESGAN received the highest score, with MLE-G achieving the second-highest result.
As mentioned in Section~\ref{armodelanalysis}, MLE-G, despite facing challenges in generating diverse sentences, was able to produce high-quality sentences.

\subsection{General Applicability}
\begin{figure*}[t]
  \begin{subfigure}{.33\textwidth}
    \centering
    \includegraphics[width=1\textwidth]{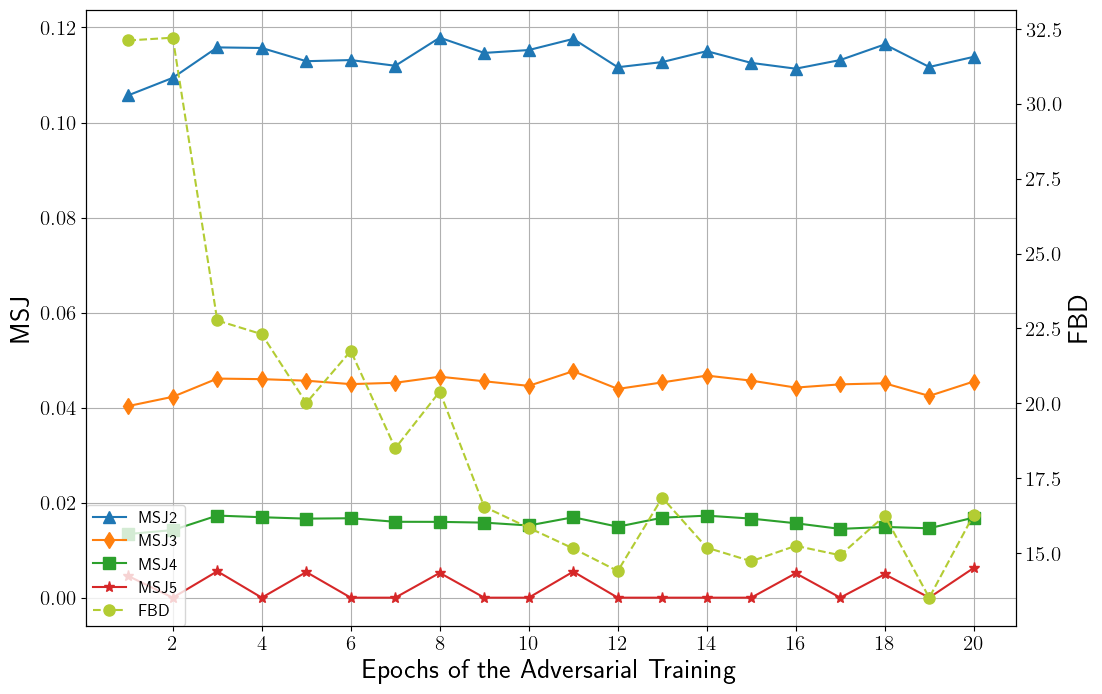}
    \caption{Zero-shot results}
    \label{zero}
  \end{subfigure}
  \begin{subfigure}{.33\textwidth}
    \centering
    \includegraphics[width=1\textwidth]{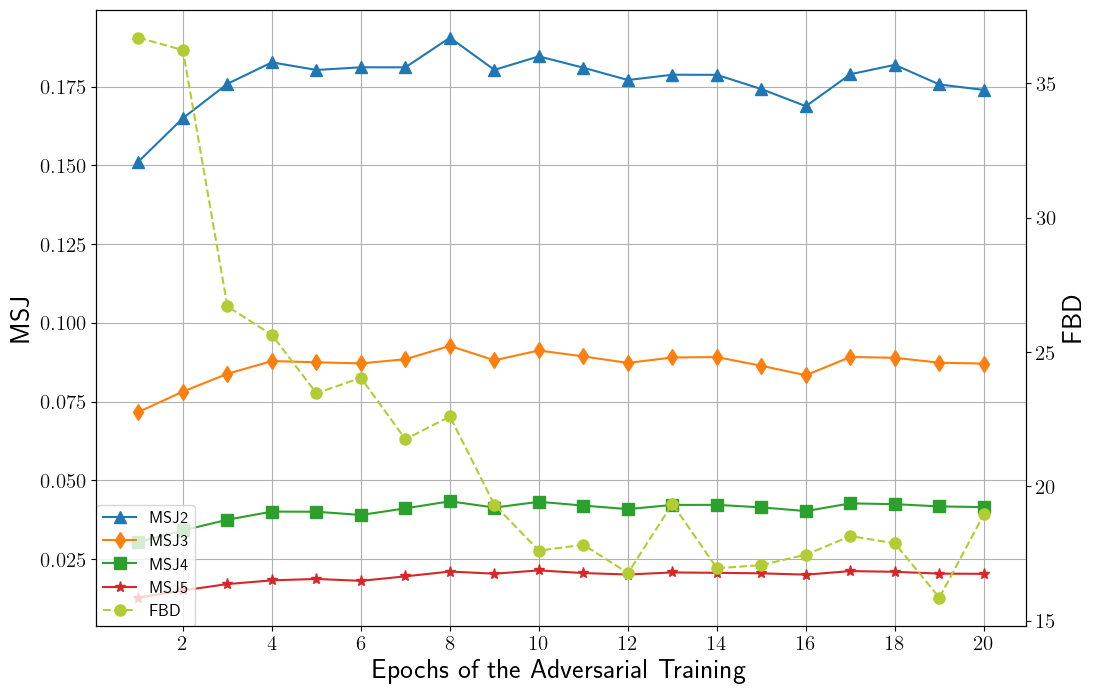}
    \caption{Non-zero-shot results}
    \label{nonzero}
  \end{subfigure}
  \begin{subfigure}{.33\textwidth}
    \centering
    \includegraphics[width=1\textwidth]{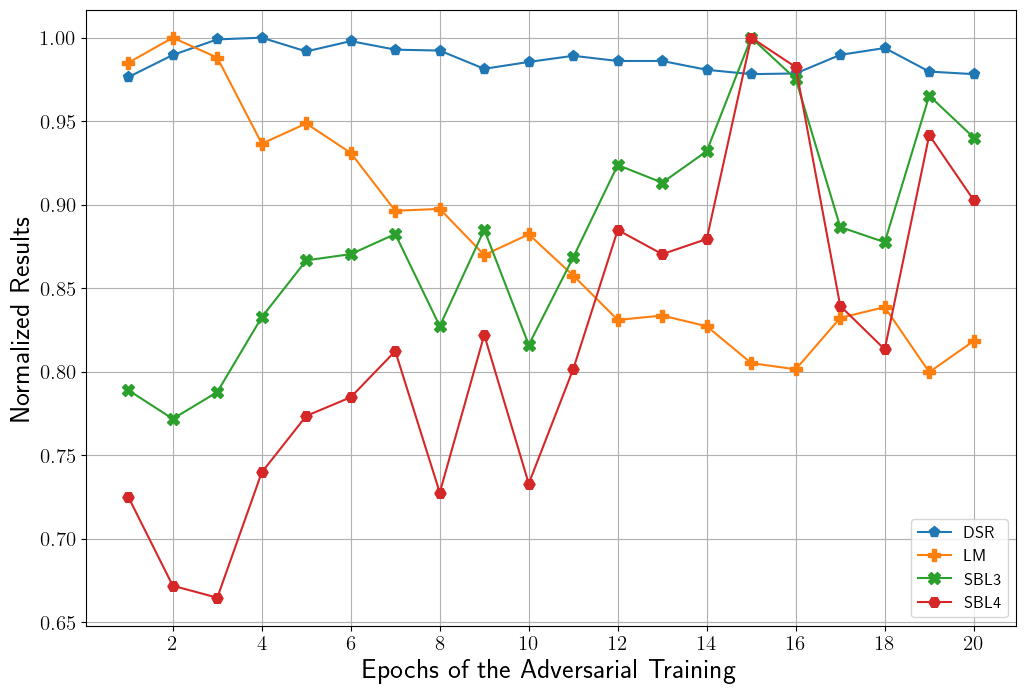}
    \caption{DSR, LM, SBL results}
    \label{imdbOthers}
  \end{subfigure}
  \caption{Zero-shot, non-zero-shot results of IMDb-trained TESGAN.
           DSR, LM, and SBL results of IMDb-trained TESGAN are normalized for ease of viewing.}
  \label{zeroshotResults}
\end{figure*}
In this section, we trained TESGAN with IMDb using a larger volume than DailyDialog to assess the general applicability of TESGAN.
The IMDb-trained TESGAN is evaluated with both the DailyDialog test set (zero-shot) and the IMDb test set (non-zero-shot).
Figures~\ref{zero} and \ref{nonzero} display the zero-shot and non-zero-shot test results of the IMDb-trained TESGAN, respectively. 
Additionally, the zero-shot results generally exhibit a similar trend to the non-zero-shot test, suggesting that the model is being trained without bias toward the training data.
Furthermore, the second group in Table~\ref{table3} presents the synthetic text results of the IMDb-trained TESGAN.
Both the zero-shot and text synthesis results indicate that TESGAN's outcomes do not vary significantly depending on the dataset, implying that TESGAN generalizes well and can be trained on diverse datasets.
Figure~\ref{imdbOthers} also illustrates the DSR, LM, and SBL results of the IMDb-trained TESGAN.
Since these metrics are evaluated not on the test set but on generated text data, they consistently yield results regardless of the zero-shot test.

\subsection{Error Analysis}
\begin{figure*}[t]
  \begin{subfigure}{.33\textwidth}
    \centering
    \includegraphics[width=1\linewidth]{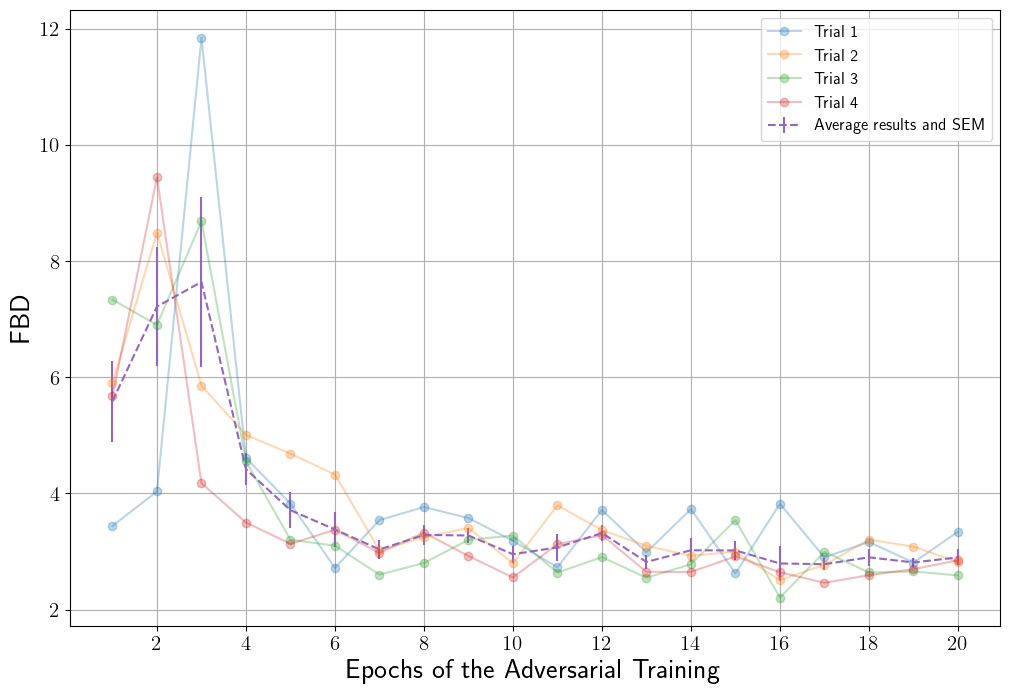}
    \caption{FBD $\downarrow$}
    \label{figure7a}
  \end{subfigure}
  \begin{subfigure}{.33\textwidth}
    \centering
    \includegraphics[width=1\linewidth]{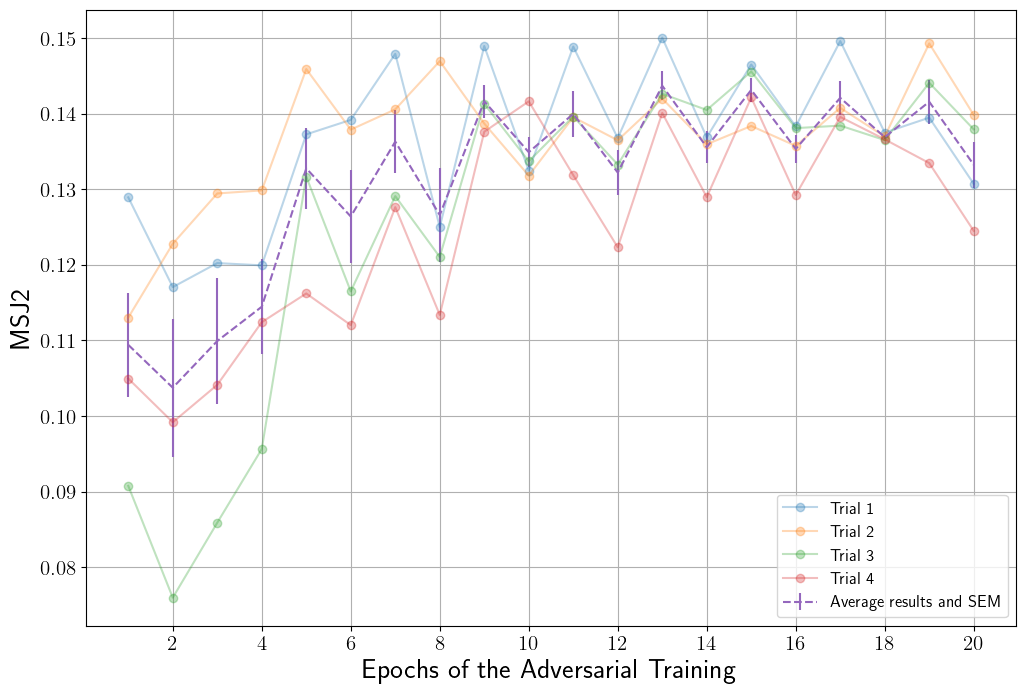}
    \caption{MS-Jaccard2 $\uparrow$}
    \label{figure7b}
  \end{subfigure}
  \begin{subfigure}{.33\textwidth}
      \centering
      \includegraphics[width=1\linewidth]{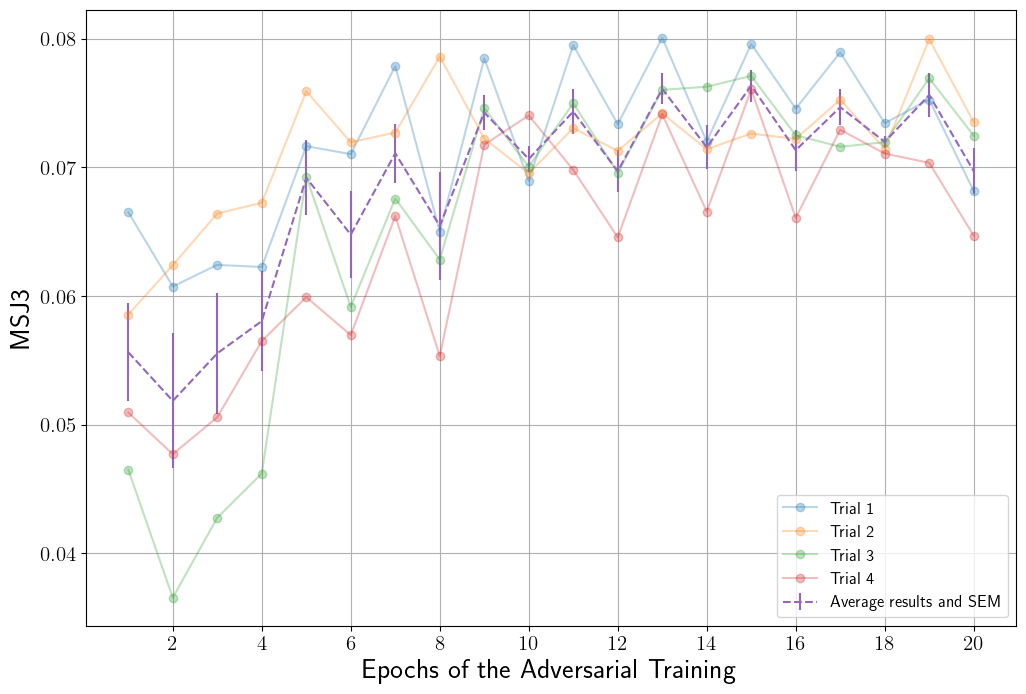}
      \caption{MS-Jaccard3 $\uparrow$}
      \label{figure7c}
  \end{subfigure}
  \begin{subfigure}{.33\textwidth}
      \centering
      \includegraphics[width=1\linewidth]{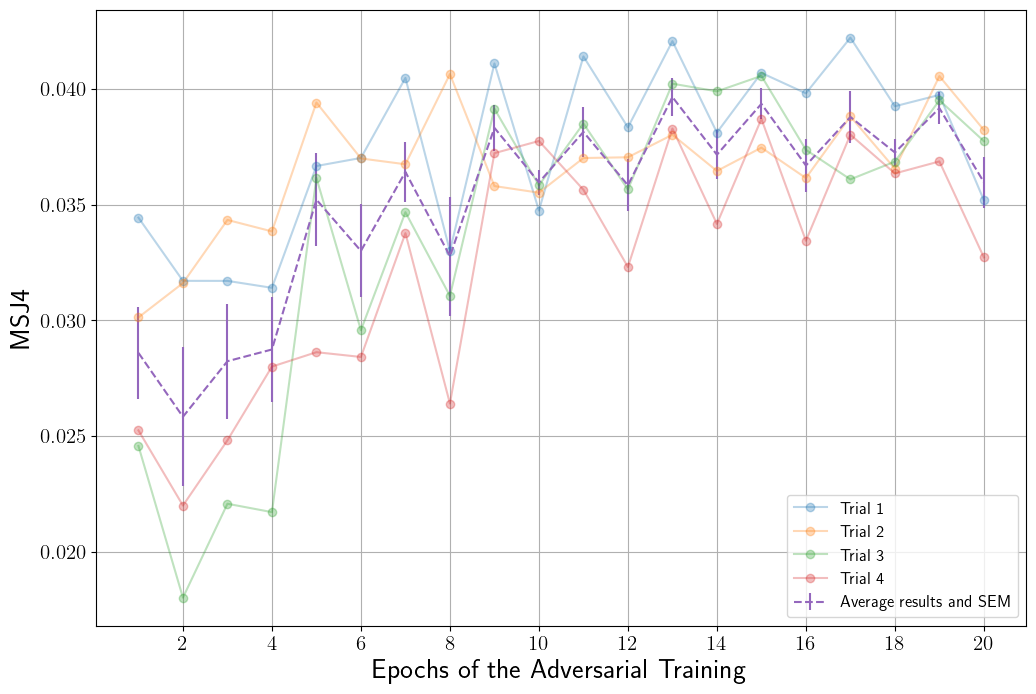}
      \caption{MS-Jaccard4 $\uparrow$}
      \label{figure7d}
    \end{subfigure}
    \begin{subfigure}{.33\textwidth}
      \centering
      \includegraphics[width=1\linewidth]{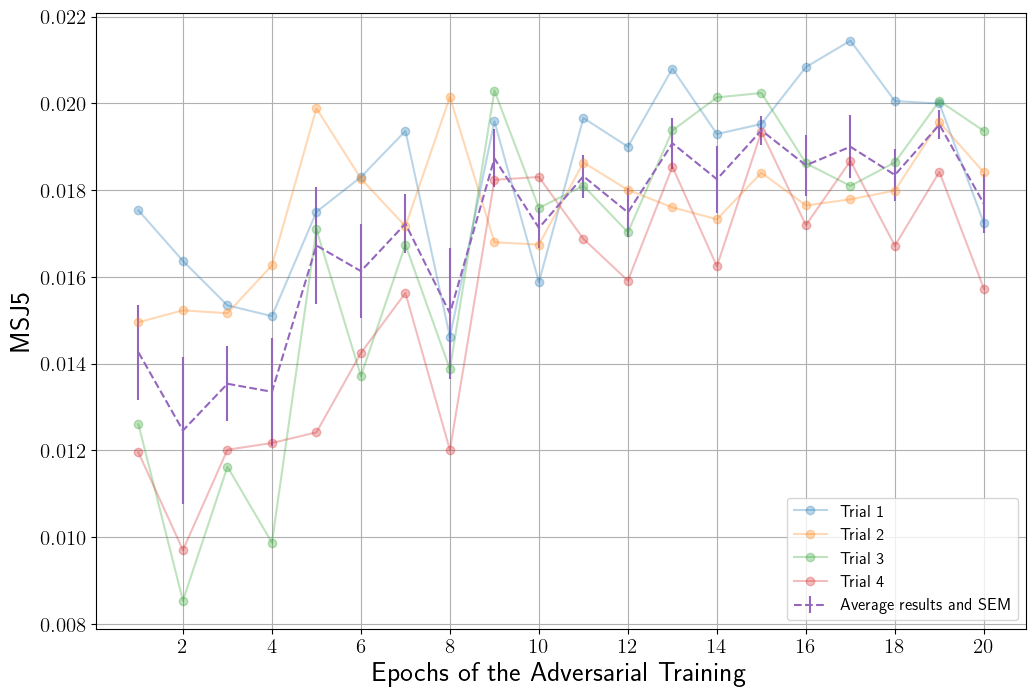}
      \caption{MS-Jaccard5 $\uparrow$}
      \label{figure7e}
    \end{subfigure}
    \begin{subfigure}{.33\textwidth}
        \centering
        \includegraphics[width=1\linewidth]{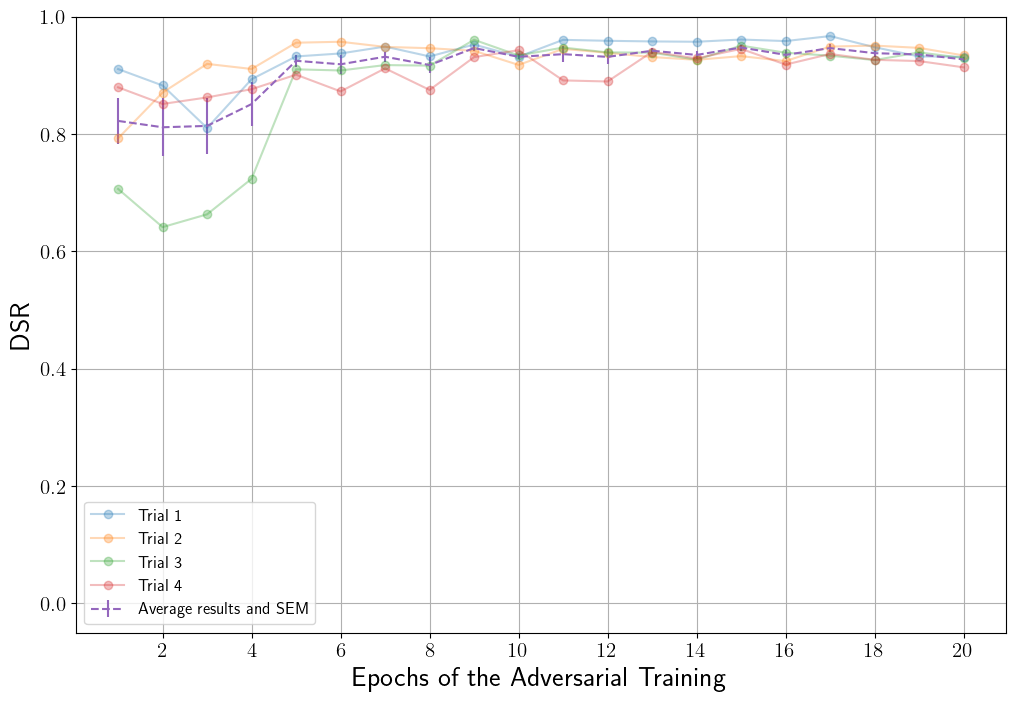}
        \caption{DSR $\uparrow$}
        \label{figure7f}
    \end{subfigure}
  \caption{Illustration of TESGAN training results. TESGANs show similar trends for every trial, and SEMs decrease during adversarial training.}
  \label{figure7}
\end{figure*}
We also conducted three additional TESGAN trainings without setting a manual seed in the code to confirm reproducibility.
To assess whether the sentences generated by the model for each trial converge as adversarial training progresses, we calculated the Standard Error of the Mean (SEM) based on the average results.
SEM is equivalent to the standard deviation of a sample mean taken from a population and represents the standard deviation that indicates the extent of variability in sample means.
SEM is calculated by $\frac{\sigma}{\sqrt{n}}$ ($\sigma$ and $n$ denote average results and the number of trials).
As a result, the overall tendency of training outcomes during adversarial training is similar.
Furthermore, the SEM of each epoch decreases during adversarial training, indicating that each TESGAN converges. 
Figure~\ref{figure7} displays the results of the four experiments, including the average results and SEM.

\section{Discussion}
\subsection{Generator and Training Strategy}
\label{generatorAndTrainingStrategy}
We found that the performance of the TESGAN framework depends on the generator's architecture.
When ReLU was used, dying ReLU~\cite{Lu2019DyingRA} occurred, where the negative values became zero, making it unsuitable for text synthesis where diversity is important.
Additionally, the hyperbolic tangent (tanh) was not adequate due to the problem of gradient vanishing~\cite{WANG201988}.
Consequently, we adopted Leaky ReLU~\cite{Maas13rectifiernonlinearities} as the activation function between two convolutional layers of the generator.
Furthermore, deep structures and batch normalization tended to result in monotonous text synthesis.
Therefore, we designed the generator's layers to be wide rather than deep without batch normalization.

We also observed that the convergence of TESGAN depends on the parameter update rate of the discriminators and the generator. 
As in Algorithm~\ref{algorithm1}, the discriminators update their parameters only during odd training epochs to allow the generator to catch up with the discriminator's learning because the convergence of the generator is commonly slower than that of the discriminator.
When the discriminators updated their parameters at every training epoch, the same as the generator, adversarial training became unbalanced.
Additionally, we conducted further experiments by changing the update frequency of the generator from once to three times per mini-batch step.
When the generator updated only once per step, the same as the discriminators, it could not keep up with the learning of the discriminators.
On the other hand, when the generator updated three times per step, the discriminators could not keep up with the learning of the generator.
Therefore, we chose to update the generator twice per step, resulting in the generator being updated four times more frequently per two epochs than the discriminators, as explained in Algorithm~\ref{algorithm1}.

\subsection{Ablation Study}
\begin{table*}[t]
  \centering
  \begin{tabular}{lccccccc}
      \hline
      Method & FBD $\downarrow$ & MSJ2 $\uparrow$ & MSJ3 $\uparrow$ & MSJ4 $\uparrow$ & MSJ5 $\uparrow$ & DSR & LM$\ast$ $\downarrow$ \\
      \hline
      TESGAN w/o SSD (13) & \textbf{2.8346} & 0.1377 & 0.0744 & 0.0394 & 0.0204 & 0.9497 & \textbf{4.1833} \\
      TESGAN w/o SOD (2) & 4.7724 & 0.1125 & 0.0596 & 0.03115 & 0.0164 & 0.8656 & 4.7011 \\
      TESGAN w/o SDP (13) & 38.1301 & 0.0580 & 0.0252 & 0.0099 & 0.0041 & 0.7390 & - \\
      TESGAN w/o SFP (16) & 2.9202 & 0.1477 & \textbf{0.0790} & \textbf{0.0422} & 0.0209 & 0.9463 & 4.2339 \\
      TESGAN (17)         & 2.8994 & \textbf{0.1496} & 0.0789 & \textbf{0.0422} & \textbf{0.0214} & \textbf{0.9669} & 4.2361 \\
      \hline
  \end{tabular}
  \caption{Results of the ablation study.
  Numbers in parentheses indicate the training epoch of the selected model.
  $\ast$ denotes a metric not considered when selecting the best model.
  - denotes that the confidence of the result is low because the quality of the synthesized sentence is poor.}
  \label{ablationTable}
\end{table*}
In this section, we confirm the effect of the four objective functions in Section~\ref{objectiveFunctions}, and the results are shown in Table~\ref{ablationTable}.
When Seed Order Discriminator (SOD) and Seed Distribution Prediction (SDP) were not used, there was a significant difference in the results, indicating that SOD and SDP are important for high-quality text synthesis.
Since MSJ evaluates text based on the n-gram of tokens, the order of the synthesized text is important.
Accordingly, the MSJ results of the "w/o SOD" in Table~\ref{ablationTable} are worse than those of the "w/o Seed Structure Discriminator (SSD)", which proves that SOD can capture the token order representations.
The four objective functions were used to achieve a good overall result, demonstrating that each of the four objective functions is playing a unique role.

\subsection{Activation Function Study}
\begin{table*}[htb!]
  \centering
  \begin{tabular}{cc|ccccccc}
    \hline
    \multicolumn{2}{c|}{Activation} & FBD $\downarrow$ & MSJ2 $\uparrow$ & MSJ3 $\uparrow$ & MSJ4 $\uparrow$ & MSJ5 $\uparrow$ & DSR $\uparrow$ & LM$\ast$ $\downarrow$ \\
    \hline
    \multirow{3}{*}{TESGAN} & None & 47.261 & 0.108 & 0.060 & 0.032 &  0.016 & \textbf{0.982} & - \\
    & Tanh & 9.780 & 0.110 & 0.057 & 0.030 & 0.015 & 0.871 & 5.402 \\
    & Sigmoid & \textbf{2.899} & \textbf{0.150} & \textbf{0.079} & \textbf{0.042} & \textbf{0.021} & 0.967 & \textbf{4.236} \\
    \hline
    \multirow{3}{*}{P-TESGAN} & None & 54.002 & 0.111 & 0.059 & 0.030 & \textbf{0.015} & \textbf{0.958} & - \\
    & Tanh & 20.158 & 0.118 & 0.061 & 0.031 & \textbf{0.015} & 0.937 & - \\
    & Sigmoid & \textbf{2.274} & \textbf{0.131} & \textbf{0.066} & \textbf{0.032} & 0.014 & 0.841 & \textbf{3.642} \\
    \hline
  \end{tabular}
  \caption{Performance according to the activation functions of the generator.
    $\ast$ denotes a metric not considered when selecting the best model.
    - denotes that the confidence of the result is low because the quality of the synthesized sentence is poor.}
    \label{table4}
\end{table*}

\begin{table*}[htb!]
      \centering
      \begin{tabular}{|l|l|}
          \hline
          \textbf{TESGAN with Tanh} & \textbf{P-TESGAN with Tanh}\\
          \hline
          You are a little & You ’ re a book?\\
          I ’ m sorry to see you off. You ’ Ve come & You are late.\\
          I ’ m sorry. & I ’ m doing ’ t “ all day ’ s\\
          You ’ d like a tour to see the dentist. & I don't know what time it is?\\
          You are late. & I ’ m sorry to hear this!\\
    
          \hline\hline
          \textbf{TESGAN without activation} & \textbf{P-TESGAN without activation}\\
          \hline
          I ’ d like to say it! I ’ d like to & I ’ s a big, that ’ s right.\\
          Yes, do you want to buy? & I like the back ones. They look like a shop.\\
          I ’ s right over there? & I ’, this ’, this ’! be real, \\
          What's the matter? & I have a problem with my English textbooks.\\
          I got a bite the food? & I ’ s faster, George. I ’ d like to go \\
          \hline
      \end{tabular}
      \caption{Synthesized sentences by tanh and non-use cases in Table~\ref{table4}. P-TESGAN denotes the perturbed TESGAN.}
    \label{table5}
\end{table*}
The results varied depending on the activation functions used at the end of the seed-making process.
We conducted experiments on sigmoid, tanh, and non-use cases during the seed-making process, and their results are shown in Table~\ref{table4}. 
Table~\ref{table5} shows the quality of the synthetic text for tanh and non-use cases, and the results are worse than those using sigmoid in Table~\ref{table3}. 
However, according to Table~\ref{table4}, the DSR results of the non-use case are higher than the sigmoid case.
Thus, we can see that a higher DSR does not always mean good quality because DSR only considers data memorization. 
Therefore, we select the model using the sigmoid activation function, which has better results for FBD and MSJ, and moderately high DSR.

\subsection{Data Memorization Study}
\begin{figure}[tb!]
    \centering
    \includegraphics[width=1\columnwidth]{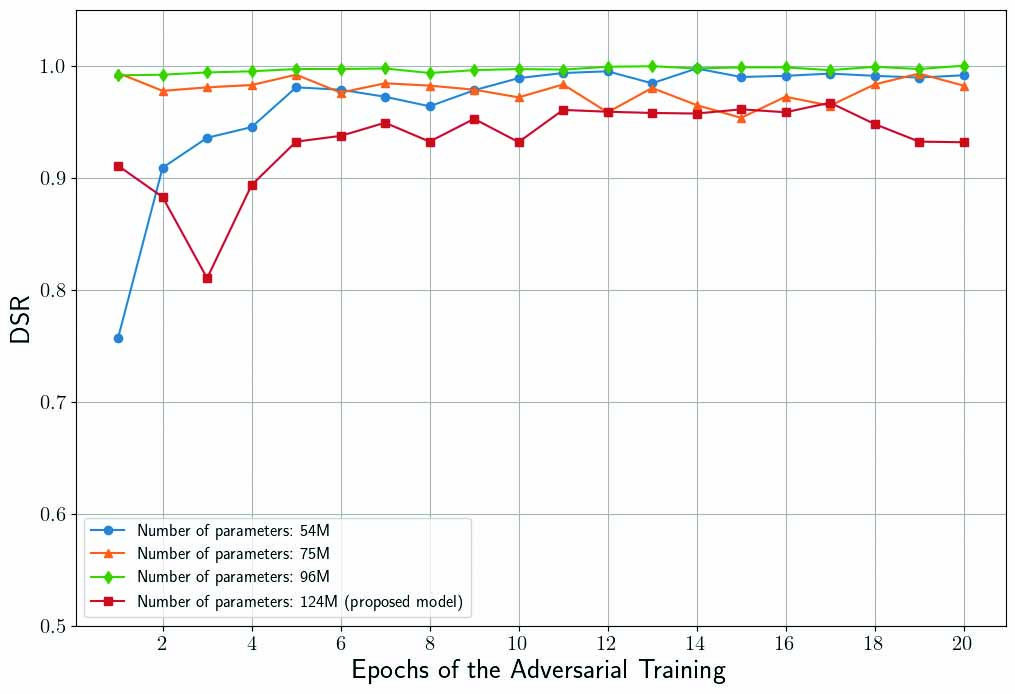}
    \caption{DSR results according to scales of seed interpretation model.}
    \label{figure6}
\end{figure}
\noindent The pre-trained GPT-2, which has 124M parameters and is used as the seed interpretation model, has been trained on relatively large corpora.
Therefore, we need to confirm whether the low data memorization comes from transfer learning or the TESGAN framework.
We trained three smaller seed interpretation models from scratch to measure the data memorization and they have 54M, 75M, and 96M parameters each.
As shown in Figure~\ref{figure6}, DSR is high regardless of the number of model parameters during adversarial training, indicating that the low data memorization comes from the TESGAN framework.

\section{Conclusion}
In this work, we proposed a novel unsupervised text synthesis framework, TESGAN.
TESGAN facilitated the gradient backpropagation of natural language discrete tokens by creating a continuous text embedding space called a seed.
In most text-GAN studies, data memorization had been inevitable because the generator had to be pre-trained with an autoregressive approach before adversarial training.
Therefore, we introduced TESGAN, which mitigated the data memorization issue by applying an unsupervised GAN framework that does not directly refer to the training data.
TESGAN improved text synthesis performance during adversarial training and resulted in the best or comparable results in terms of evaluation metrics. 
Additionally, TESGAN exhibited the lowest data memorization ratio, and the data memorization study confirmed that these results were attributable to the TESGAN framework. 
Furthermore, TESGAN achieved the highest scores in human evaluations.
The ablation study highlighted the importance of the four objective functions, and the synthetic text results from a large dataset-trained TESGAN demonstrated its general applicability. 
This paper underscores the potential of continuous embedding spaces in conjunction with discrete tokens for text synthesis through unsupervised learning. 
By integrating the concept of viewing text as a continuous space with publicly available Large Language Models \cite{touvron2023llama}, models can synthesize more expressive sentences, and we anticipate that many follow-up studies will emerge.

\clearpage
\bibliography{nejlt}
\bibliographystyle{nejlt_bib}
\clearpage

\appendix
\section{Details of Models}
\label{appendixA}

\subsection{Seed Interpretation Model}
The seed interpretation model $f_{\theta}(\cdot)$ is necessary to predict subsequent sentences a seed makes.
Therefore, the seed interpretation model must be trained with multi-turn sentences in an autoregressive way.
Our interpretation model inherits the 12-layer GPT-2, derived from the decoder of the transformer language model \cite{NIPS2017_3f5ee243}, and has 124M parameters.
We used the model achieving the highest NLTK BLEU-4\footnote{\url{https://www.nltk.org/_modules/nltk/translate/bleu_score.html}} in the validation set of each dataset in Table~\ref{table1} as the seed interpretation model.

\subsection{Generator}
The generator $g_{\phi}(\cdot)$ consists of two 1D transposed convolutional layers and has 3.3M parameters.
The first and the second layers conduct convolution with 128 and 16 filters, respectively.
Since sentences vary according to types of tokens and their order, forms of the real seed $H_{real}$ are also varied.
The generator generates seeds from the uniform distribution noise $X$ with an interval of $[-10,10)$ to make diverse forms of the seeds.
Furthermore, the fake seeds generated by the deep convolutional layers and batch normalization results tend to synthesize only monotonous sentences.
Thus, layers of the generator are constructed not deeply but widely and the generator does not have batch normalization layers.
Leaky ReLU is used as the activation function between the two convolutional layers.

\subsection{Seed Structure Discriminator (SSD)}
Sentence structure is important for constructing a complete sentence.
For example, “\textit{I love you so much}” is structurally error-free, but “\textit{I love like so much}” and “\textit{I love}” are not.
Because real seeds are created from perfect sentences, they retain the structural representation of sentences.
Therefore it is important that the fake seeds should capture the structural features of the real seeds.
We assume that every sentence can be the first sentence in multi-turn cases.
Thus, the real seeds are obtained from sentences where the $[CLS]$ token is inserted at the beginning like Equation~\ref{equation2}.
We use the 2-layer BERT $d_{\alpha}(\cdot)$ to capture the structural features of sentences, and the $[CLS]$ token's feature is used to predict whether the seed is real (label 1) or fake (label 0).
In addition, real and fake seeds do not pass through the embedding part of the BERT because they are already embedding spaces.
Finally, the BERT used in SSD has 54M parameters.

\subsection{Seed Order Discriminator (SOD)}
The order of tokens is important for constructing sentences.
For example, “\textit{I love you so much}” is syntactically correct, but “\textit{I you love so much}” and “\textit{I love you much so}” are not.
We use a 2-layer Bidirectional LSTM to consider both forward and backward directions of sentences and the model has 24M parameters.
The concatenated hidden states of the last token ($[SEP]$ or $[PAD]$) and the first token ($[CLS]$) are used to predict whether the seed is real (label 1) or fake (label 0).

\section{Loss Function}
\label{losses}
Here, we show whole loss functions of TESGAN:
\scriptsize
\begin{multline}
\label{allLosses}
  \textit{Seed Interpretation Model}: \\ 
    \mathcal{L}_{LM}=-\frac{1}{N}\sum^{N}_{n=1}{\log{\frac{exp(x_{n, y_n})}{\sum^{C}_{c=1}{exp(x_{n,c})}}}} \\\\
  \textit{Adversarial Training Training} \\
    \mathrm{d}-\mathrm{step}: \\
    \mathcal{L}_{D} = -\frac{1}{N}\sum_{i=1}^{N} \Big\{ \big[y_i\log{x_i} + (1-y_i) \log{(1 - x_i)} \big]_{SSD}^{real/fake} + \\
                          \big[y_i\log{x_i} + (1-y_i) \log{(1 - x_i)} \big]_{SOD}^{real/fake} \Big\} \\\\
    \mathrm{g}-\mathrm{step}: \\
    \mathcal{L}_{G} = -\frac{1}{N}\sum_{i=1}^{N} \Big\{ \big[y_i\log{x_i} + (1-y_i) \log{(1 - x_i)} \big]_{SSD}^{fake} + \\
                          \big[y_i\log{x_i} + (1-y_i) \log{(1 - x_i)} \big]_{SOD}^{fake} + \\
                          \sigma\big(f_{\theta}(H_{real})\big)\log{\frac{\sigma\big(f_{\theta}(H_{real})\big)}{\sigma\big(f_{\theta}(H_{fake})\big)}} +\\
                          \|\mu_{r} - \mu_{f}\|_2^2 + \|H_{real} - H_{fake}\|_1 \Big\} \\
\end{multline}
\normalsize
The first loss function in Equation~\ref{allLosses} is cross-entropy and is used to train the seed interpretation model.
The loss functions used in adversarial training operate differently in the discriminator and generator steps.
In the discriminator step (d-step), the loss function is designed to train the discriminator to distinguish between real and fake seeds, predicting them as 1 and 0, respectively.
On the other hand, in the generator step (g-step), the loss function aims to train the generator to predict fake seeds as 1.
Additionally, SDP and SFP losses are added to assist the generator learning during the g-step.

\section{Statistics of Datasets}
\label{datasetStats}
Table~\ref{table1} shows the statistics of the two datasets used in this paper.
We excluded single-turn reviews when constructing the IMDb multi-turn dataset.
For the baseline performance experiments, we generated fake seeds equal to the number of sentences in the DailyDialog dataset to conduct TESGAN training (adversarial training).
Similarly, for the general applicability experiments, we generated nearly 300k fake seeds to conduct the experiments with IMDb datasets.
\begin{table*}[htb!]
  \centering
  \begin{tabular}{lrrrrrr}
      \hline
      \multirow{2}{*}{Statistics} & \multicolumn{3}{c}{DailyDialog} & \multicolumn{3}{c}{IMDb} \\
                            & Train & Validation & Test & Train & Validation & Test \\
      \hline
      $\#$ of multi-turn set & 11,118 & 1,000 & 1,000 & 24,890 & 12,500 & 12,390 \\
      Total sentences & 87,170 & 8,069 & 7,740 & 299,137 & 150,369 & 148,768 \\
      Avg. $\#$ of turns per set & 7.84 & 8.07 & 7.74 & 12.02 & 12.03 & 12.01 \\
      Avg. $\#$ of words per sentence & 11.30 & 11.21 & 11.44 & 19.34 & 19.40 & 19.28 \\
      Avg. $\#$ of tokens per sentence & 14.51 & 14.39 & 14.69 & 24.25 & 24.31 & 24.20 \\
      \hline
  \end{tabular}
  \caption{Statistics of datasets}
  \label{table1}
\end{table*}

\section{Hyperparameters}
\label{hyperparameters}
The TESGAN framework has two training steps.
The first step is seed interpretation model training.
The multi-turn data for seed interpretation model training were limited to a maximum of four and eight turns in performance (DailyDialog-trained) and general applicability (IMDb-trained) experiments, respectively.
Also, the maximum length of the sentence was set to 16 and 32 for each experiment (total sequence length of each experiment was 64 and 256).
The 12-layer GPT-2 is used as the seed interpretation model, and both hidden and embedding dimensions are 768.
We adopted the byte-pair-encodings (BPE) \cite{sennrich-etal-2016-neural} tokenizer with 50,260 vocabularies in the seed interpretation model.
We used the Adam optimizer \cite{kingma2014method} with $1e^{-3}$ learning rate to train the seed interpretation model and set the mini-batch size to 100.

In the adversarial training phase, the sentences used as the seeds are composed of tokens explained in Equation~\ref{equation2}, and the length of each sentence is set to 16 including special tokens.
The discriminators are updated by the loss of SSD and SOD during adversarial training.
Also, the generator is updated not only by the loss of SSD and SOD but also that of SDP and SFP.

The fake seeds are generated from the uniform distribution noise $X$ with an interval of $[-10,10)$ by the generator, which has two convolutional processes.
In addition, the Leaky ReLU with slope 0.5 and the sigmoid are used in the middle and the end of the generator, respectively.
We used the Adam optimizer with $2e^{-4}$ learning rate when training DailyDialog because the generator has difficulty converging when the learning rate exceeds $4e^{-4}$.
However, when training on IMDb, a larger dataset than DailyDialog, we set the learning rate to $5e^{-4}$.
The BERT and the LSTM models, used as SSD and SOD respectively, consist of two layers and 768 hidden dimensions.
Both discriminators used the Adam optimizer with $5e^{-4}$ and $1e^{-3}$ learning rate, respectively.
When the learning rates of the discriminators were larger than the proposed values, adversarial learning was imbalanced.
Also, the mini-batch size was set to 128 during adversarial training.
Lastly, all the above experiments took place on a machine with Ubuntu 18.04.5 and an NVIDIA RTX 3090 GPU.

\section{Additional Results}
\label{baselinesPerformanceResults}
\begin{figure*}[htb!]
  \begin{subfigure}{.33\textwidth}
    \centering
    \includegraphics[width=1\linewidth]{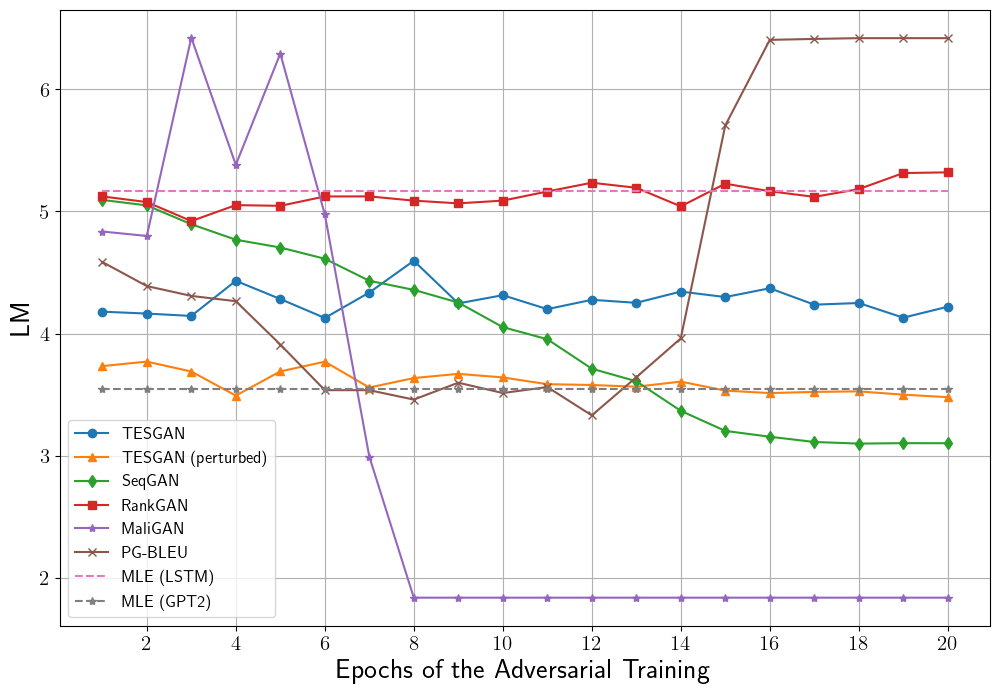}
    \caption{LM $\downarrow$}
    \label{figure9a}
  \end{subfigure}
  \begin{subfigure}{.33\textwidth}
    \centering
    \includegraphics[width=1\linewidth]{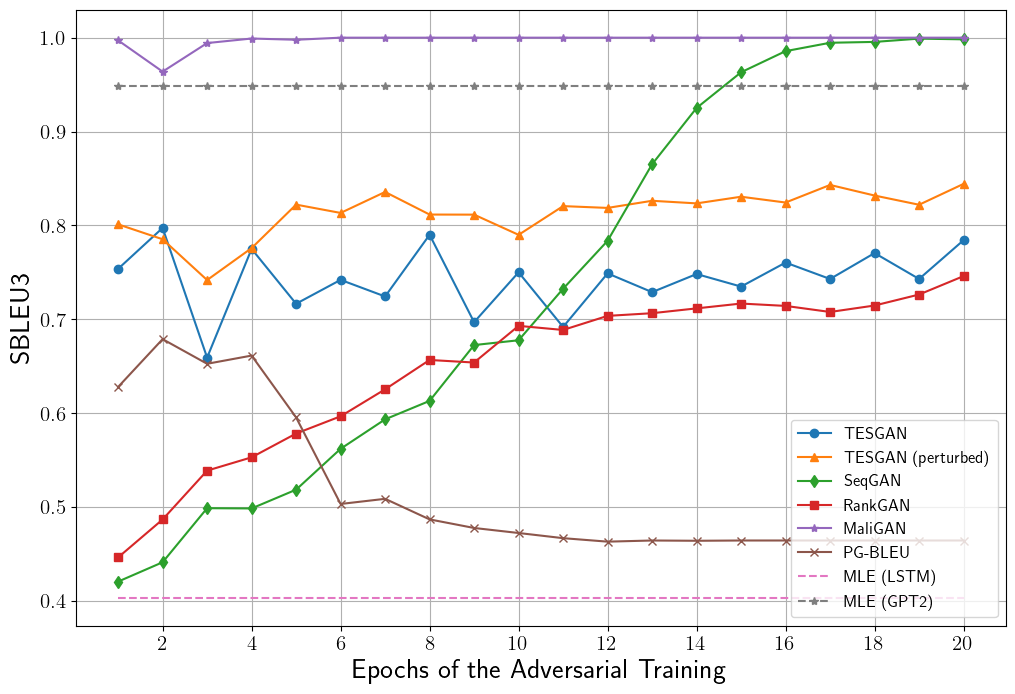}
    \caption{SBL3 $\downarrow$}
    \label{figure9b}
  \end{subfigure}
  \begin{subfigure}{.33\textwidth}
      \centering
      \includegraphics[width=1\linewidth]{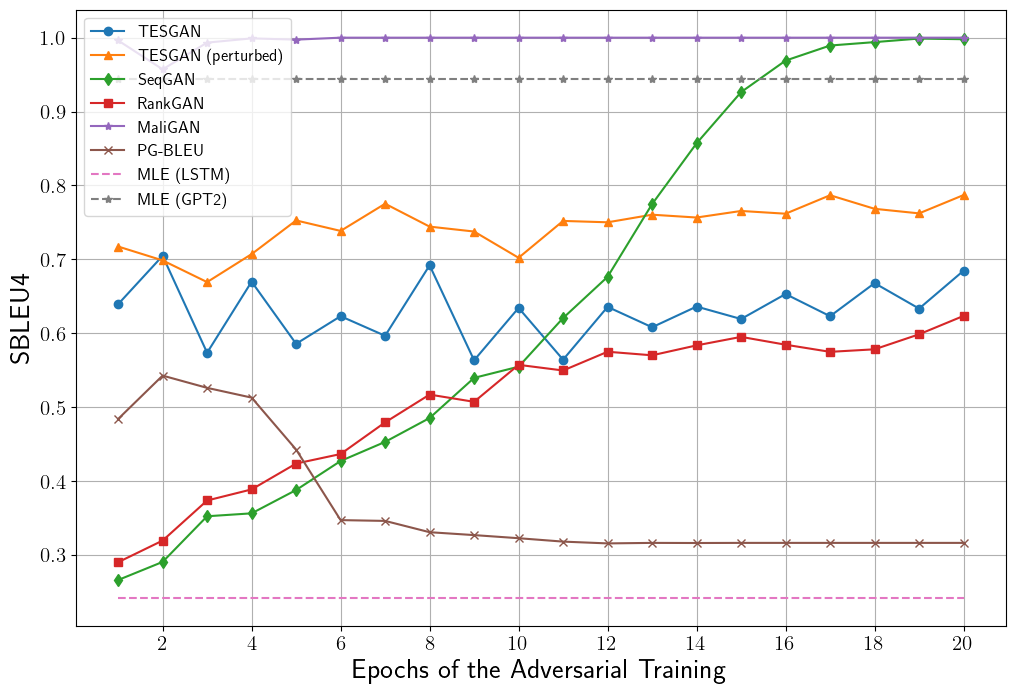}
      \caption{SBL4 $\downarrow$}
      \label{figure9c}
  \end{subfigure}
  \caption{LM, SBL results of TESGAN-based models and baselines trained with DailyDialog.}
  \label{figure9}
\end{figure*}

We provide evaluation results of the text generated by each model per epoch.
Table~\ref{epochResults} shows the results of 1, 5, 10, 15, 17, and 20-epoch results of each model.
Also, Figure~\ref{figure9} shows LM and SBL results of the TESGAN-based models and the baselines.
In Figure~\ref{figure9}, the SBL results of the baselines tend to increase.

\begin{table*}[htb!]
  \centering
  \scriptsize
  \begin{tabular}{llrrrrrr}
      \hline
      Method & Epoch & FBD $\downarrow$ & MSJ2 $\uparrow$ & MSJ3 $\uparrow$ & MSJ4 $\uparrow$ & MSJ5 $\uparrow$ & DSR ($R_{syn}, R_{unq}$) $\uparrow$ \\
      \hline
      \multirow{5}{*}{TESGAN} & 1  & 3.441 & 0.129 & 0.067 & 0.034 & 0.018 & 0.911 (1, 0.836) \\
                              & 5  & 3.826 & 0.137 & 0.072 & 0.037 & 0.018 & 0.932 (1, 0.873) \\
                              & 10 & 3.185 & 0.132 & 0.069 & 0.035 & 0.016 & 0.932 (0.999, 0.873) \\
                              & 15 & \textbf{2.624} & 0.146 & \textbf{0.080} & 0.041 & 0.020 & 0.961 (1, 0.925) \\
                              & 17 & 2.899 & \textbf{0.150} & 0.079 & \textbf{0.042} & \textbf{0.021} & \textbf{0.967} (1, 0.936) \\
                              & 20 & 3.339 & 0.131 & 0.068 & 0.035 & 0.017 & 0.932 (1, 0.872) \\
      \hline
      \multirow{5}{*}{P-TESGAN} & 1  & 6.146 & 0.112 & 0.057 & 0.029 & 0.013 & 0.803 (1, 0.671) \\
                                & 5  & 3.746 & 0.118 & 0.060 & 0.030 & \textbf{0.016} & 0.801 (0.998, 0.669) \\
                                & 10 & 2.274 & \textbf{0.131} & \textbf{0.066} & \textbf{0.032} & 0.014 & \textbf{0.841} (0.997, 0.727) \\
                                & 15 & \textbf{2.132} & 0.121 & 0.061 & 0.030 & 0.014 & 0.812 (1, 0.683) \\
                                & 17 & 2.274 & 0.122 & 0.063 & 0.031 & 0.014 & 0.789 (0.998, 0.653) \\
                                & 20 & 2.309 & 0.119 & 0.058 & 0.029 & 0.014 & 0.784 (0.997, 0.646) \\
      \hline
      \multirow{5}{*}{SeqGAN} & 1  & \textbf{6.153} & \textbf{0.185} & \textbf{0.091} & \textbf{0.040} & \textbf{0.015} & \textbf{0.880} (0.883, 0.877) \\
                              & 5  & 6.373 & 0.121 & 0.065 & 0.032 & 0.014 & 0.602 (0.639, 0.568) \\
                              & 10 & 9.432 & 0.064 & 0.034 & 0.017 & 0.007 & 0.289 (0.34, 0.251) \\
                              & 15 & 18.471 & 0.012 & 0.006 & 0.003 & 0 & 0.019 (0.025, 0.015) \\
                              & 17 & 24.504 & 0.004 & 0.002 & 0.001 & 0 & 0.004 (0.006, 0.003) \\
                              & 20 & 28.048 & 0 & 0 & 0 & 0 & 0.001 (0.013, 0.001) \\
      
       \hline
      \multirow{5}{*}{RankGAN} & 1  & \textbf{6.409} & \textbf{0.189} & \textbf{0.096} & \textbf{0.048} & \textbf{0.023} & \textbf{0.890} (0.895, 0.886) \\
                               & 5  & 6.778 & 0.160 & 0.084 & 0.04 & 0.016 & 0.82 (0.851, 0.791) \\
                               & 10 & 10.862 & 0.138 & 0.074 & 0.037 & 0.018 & 0.739 (0.799, 0.687) \\
                               & 15 & 9.732 & 0.118 & 0.060 & 0.030 & 0.015 & 0.699 (0.791, 0.625) \\
                               & 17 & 9.893 & 0.115 & 0.058 & 0.028 & 0.012 & 0.696 (0.792, 0.62) \\
                               & 20 & 12 & 0.114 & 0.056 & 0.026 & 0.011 & 0.721 (0.836, 0.634) \\
      \hline
      \multirow{5}{*}{MaliGAN}  & 1  & 21.436 & \textbf{0.015} & \textbf{0.006} & \textbf{0.003} & 0 & \textbf{0.030} (1, 0.015) \\
                                & 5  & \textbf{16.589} & 0.003 & 0 & 0 & 0 & 0.027 (1, 0.014) \\
                                & 10 & 57.769 & 0 & 0 & 0 & 0 & 0 (1, 0) \\
                                & 15 & 57.769 & 0 & 0 & 0 & 0 & 0 (1, 0) \\
                                & 17 & 57.769 & 0 & 0 & 0 & 0 & 0 (1, 0) \\
                                & 20 & 57.769 & 0 & 0 & 0 & 0 & 0 (1, 0) \\
      \hline
      \multirow{5}{*}{PG-BLEU}  & 1  & \textbf{9.002} & \textbf{0.071} & \textbf{0.036} & \textbf{0.015} & \textbf{0.006} & \textbf{0.569} (0.555, 0.584) \\
                                & 5  & 18.974 & 0.017 & 0.008 & 0 & 0 & 0.139 (0.472, 0.082) \\
                                & 10 & 21.568 & 0.001 & 0.001 & 0 & 0 & 0.041 (0.792, 0.021) \\
                                & 15 & 89.764 & 0 & 0 & 0 & 0 & 0.004 (0.773, 0.002) \\
                                & 17 & 142.384 & 0 & 0 & 0 & 0 & 0.003 (1, 0.001) \\
                                & 20 & 142.384 & 0 & 0 & 0 & 0 & 0.001 (1, 0.001) \\
      \hline
  \end{tabular}
  \caption{Performance of each model per epoch.}
  \label{epochResults}
\end{table*}

\end{document}